\documentclass[11pt]{humain-lab}

\usepackage{alltt}
\usepackage{amsmath,amssymb,bm}
\usepackage{amsthm}
\usepackage{algorithm}
\usepackage[noend]{algpseudocode}

\usepackage{cleveref}
\Crefname{appendix}{Appendix}{Appendices}
\Crefname{prompt}{Prompt}{Prompts}

\usepackage{fvextra}
\usepackage{upquote}

\newcommand{\promptsize}{\fontsize{8}{9.5}\selectfont}

\newcounter{aiprompt}
\newtcolorbox{aipromptboxinner}[1]{
  title={Prompt~\theaiprompt: #1}
}
\newenvironment{aipromptbox}[1][]
  {\refstepcounter{aiprompt}\begin{aipromptboxinner}{#1}}
  {\end{aipromptboxinner}}

\title{Decision-Focused Reinforcement Learning \\
for LLM Post-Training}

\author[1,\dagger]{Haotian Zhai}
\author[1,\dagger]{Liu Leqi}

\affiliation[1]{The University of Texas at Austin}

\contribution[\dagger]{Corresponding author}

\abstract{
Large language models (LLMs) have shown promise as
predictors across various tasks, benefiting from their
ability to use information from unstructured
context.
However, when predictions serve as parameters of downstream
optimization problems, higher prediction accuracy does not
necessarily translate into lower decision costs. 
We propose decision-focused reinforcement learning (DFRL),
a framework for post-training LLMs to predict optimization
parameters using negative
downstream decision regret as reward. DFRL trains LLM predictors to minimize the costs of the
decisions induced by their predictions, without requiring
downstream convexity or differentiability, and integrates
readily into existing RL post-training frameworks. 
To address expensive reward evaluation, we develop
two approaches.
Heuristic warm-starting uses inexpensive approximate
rewards during an initial phase before switching
to refinement with exact rewards.
Reward correction combines inexpensive estimates
with randomized exact evaluations, preserving the
expected policy gradient while reducing the expected
number of exact solves per update.
Experiments on three synthetic optimization problems
demonstrate improvements over prediction-loss
baselines, while a larger knapsack experiment shows
faster early improvement under short training budgets.
On inventory planning with H\&M transactions and
product descriptions, our best configuration,
in which the LLM reasons before predicting, reduces regret by 31.4\% relative to the SFT
baseline and 12.4\% relative to the strongest
evaluated conventional forecasting baseline.
}

\correspondence{Haotian Zhai at \email{haotian.zhai@utexas.edu}, Liu Leqi at \email{leqiliu@utexas.edu}}
\reportdate{November 2026}

\begin{document}

\maketitle

\section{Introduction} \label{sec:Introduction}

Large language models (LLMs) have demonstrated strong predictive
capabilities across a range of tasks, including tabular classification~\citep{hegselmann2023tabllm,gardner2024large}, time series forecasting~\citep{gruver2023large,jin2024time}, and real-world event probability estimation~\citep{halawi2024approaching}. LLMs are particularly effective at extracting predictive
information from unstructured text, such as news and company
reports, which is difficult to represent as numerical features. For example,
\citet{wang2024news} demonstrate that incorporating relevant
news into LLM-based forecasting improves predictive accuracy.

However, in decision-making tasks where predictions serve as
parameters of downstream optimization problems, predictive
accuracy does not necessarily translate into decision quality.
Predictive models are typically trained to minimize prediction
losses, such as squared error for numerical predictions or
token-level cross-entropy for LLMs trained through supervised
fine-tuning (SFT)~\citep{ouyang2022training,touvron2023llama,
grattafiori2024llama}.
Although these objectives encourage accurate parameter predictions,
they do not explicitly account for the costs of the decisions
made using those predictions.

For example, suppose a retailer orders a product based on an
LLM's demand forecast, while actual demand is 100 units. Forecasts of 80
and 120 units have equal squared error, but with shortage
and excess inventory costs of \$10 and \$1 per unit, they
incur costs of \$200 and \$20, respectively. This difference
in decision costs is not explicitly captured by the prediction loss.

A common approach to this challenge is
decision-focused learning, which trains predictive models
to minimize the decision loss induced by their predictions
~\citep{sadana2025survey,wilder2019melding}. A fundamental challenge of decision-focused learning is differentiating through the optimizer: the Jacobian of its solution with respect to the predicted parameters can be difficult to compute, undefined, or uninformative, hindering gradient-based training. Existing methods address this challenge through implicit
differentiation~\citep{donti2017task} or surrogate
losses~\citep{elmachtoub2022smart}, but rely on structural
assumptions that limit their applicability.

Meanwhile, reinforcement learning (RL) has been successfully
applied to decision-making tasks
~\citep{bello2016neural,silver2016mastering} and, more recently,
to LLM post-training
~\citep{ouyang2022training,guo2025deepseek}.
RL can train models using reward feedback, even when the process that produces the reward is
non-differentiable~\citep{williams1992simple}.
Advances in RL post-training therefore provide practical methods
for adapting LLM predictors using feedback from downstream decisions.

Motivated by these developments, we introduce
\textbf{decision-focused reinforcement learning (DFRL)},
a framework for post-training LLMs using downstream decision
regret as reward feedback. An LLM generates predictions of unknown optimization parameters
from the available context, and a downstream optimizer selects
an optimal decision under those predictions.
We evaluate this decision under the observed parameters and
measure its regret relative to the optimal decision that could
have been made with those parameters known in advance.
The negative regret is the RL reward, so training never differentiates through the optimizer.
DFRL separates this reward formulation from the choice of RL
algorithm, allowing different policy optimization methods to
train the predictor within the same framework.

Reward evaluation can be a computational bottleneck
in DFRL because each sampled prediction requires
solving a downstream optimization problem.
We develop two approaches that use
inexpensive reward estimates to enable more policy
updates within a fixed training budget.
Heuristic warm-starting accepts biased feedback
during an initial phase, then switches to
exact-reward training for
refinement.
Reward correction combines inexpensive estimates
with randomized exact evaluations to preserve
the expected policy gradient.
We analyze when the computational savings are beneficial and evaluate
both approaches experimentally.

Our contributions are as follows:
\begin{enumerate}
    \item \textbf{Decision-focused post-training.}
    We propose \textbf{decision-focused reinforcement learning (DFRL)}, a general decision-focused learning framework for post-training LLMs to predict parameters used in downstream optimization, with negative regret as reward. DFRL integrates into
    existing RL training pipelines without prescribing a particular
    RL algorithm. 
    It requires neither differentiation through the optimizer nor convexity, linearity, or differentiability of the downstream optimization problem. 

    \item \textbf{Efficient training with inexpensive feedback.}
    We develop heuristic warm-starting and reward correction to
    reduce the cost of reward evaluation.
    Warm-starting accepts a temporary gradient bias in exchange
    for cheaper early updates, while reward correction stays
    unbiased at the cost of added variance.
    We analyze when each trade-off pays off.

    \item \textbf{Empirical results.}
    We demonstrate the effectiveness of DFRL on three classes of synthetic problems, as well as an LLM-based inventory planning task. DFRL achieves the lowest mean regret across the synthetic benchmarks. In inventory planning, it reduces regret by up to 31.4\% relative to SFT and 12.4\% relative to the strongest conventional forecasting baseline.
\end{enumerate}

\section{Background and Related Work} \label{sec:Background}

We consider decision-making problems with unknown optimization parameters. Let $x\in\mathcal X$ denote the context available before a decision is made, which may include structured and unstructured data, and let $c\in\mathcal C\subseteq\mathbb R^d$ denote the unknown parameters. A predictive model $\pi_\theta(\cdot\mid x)$, for example an LLM,  parameterized by $\theta$, specifies a conditional distribution over parameter predictions, from which we sample
\[
    \hat c \sim \pi_\theta(\cdot\mid x).
\]

Consider an optimization problem with decision variables $w\in\mathbb R^m$, feasible set $\mathcal W\subseteq\mathbb R^m$, and objective function $F(w,c)$. Since $c$ is unknown at decision time, we substitute the prediction $\hat c$ and solve
\begin{equation}
    w^\star(\hat c)
    \in
    \arg\min_{w\in\mathcal W} F(w,\hat c).
    \label{eq:induced_decision}
\end{equation}
We assume that an optimal solution exists and use a fixed tie-breaking rule when it is not unique. We evaluate the solution $w^\star(\hat c)$ under the observed parameters $c$, which are revealed after the decision is made, and define its regret as
\begin{equation}
    \ell(\hat c,c)
    :=
    F(w^\star(\hat c),c)
    -
    F(w^\star(c),c),
    \label{eq:regret}
\end{equation}
where $w^\star(c)\in\arg\min_{w\in\mathcal W}F(w,c)$ is an optimal solution under the observed parameters. 
Since $w^\star(\hat c)\in\mathcal W$ and $w^\star(c)$ minimizes $F(w,c)$ over $\mathcal W$, we have $\ell(\hat c,c)\geq 0$. 
Regret measures the additional cost incurred by solving the optimization problem using predicted rather than observed parameters.

Let $\mathcal D$ denote the joint distribution of contexts
and observed parameters $(x,c)$. The decision-focused learning
objective is to minimize expected regret:
\begin{equation}
    \min_\theta 
    \mathcal L(\theta)
    :=
    \mathbb E_{(x,c)\sim\mathcal D}
    \left[
        \mathbb E_{\hat c\sim\pi_\theta(\cdot\mid x)}
        \big[\ell(\hat c,c)\big]
    \right].
    \label{eq:dfl_objective}
\end{equation}

\subsection{Decision-Focused Learning} \label{sec: decision-focused learning}

Unlike conventional predict-then-optimize, which trains predictive models using prediction losses such as squared prediction error $\ell_{\mathrm{pred}}(\hat c,c)=\|\hat c-c\|_2^2$ and then uses their predictions as inputs to the optimization problem, decision-focused learning trains predictors according to the quality of the downstream decisions induced by their predictions. However, backpropagating the decision loss through the optimizer
presents a differentiation challenge. For a differentiable
deterministic predictor $\hat c=f_\theta(x)$, by the chain rule,
\begin{equation*}
    \nabla_\theta \ell(f_\theta(x),c)
    =
    \left(\frac{\partial f_\theta(x)}{\partial\theta}\right)^\top
    \left(\frac{\partial w^\star(\hat c)}{\partial\hat c}\right)^\top
    \left.\nabla_w F(w,c)\right|_{w=w^\star(\hat c)}.
    \label{eq:dfl_chain_rule}
\end{equation*}
The optimizer Jacobian $\frac{\partial w^\star(\hat c)}{\partial\hat c}$ can be difficult to
compute, undefined, or uninformative. For example, in mixed-integer linear programs (MILP) with
predicted objective coefficients, the optimal solution map
can be piecewise constant, yielding zero gradients almost
everywhere and becoming discontinuous where the optimal
solution changes.

Existing methods address this challenge through differentiable
optimization, smoothing, and surrogate losses.
For convex problems satisfying suitable regularity conditions,
implicit differentiation of optimality conditions enables
backpropagation through optimization
~\citep{amos2017optnet,donti2017task,agrawal2019differentiable}.
For linear and integer linear programs, continuous relaxations
combined with quadratic or log-barrier regularization provide
differentiable approximations
~\citep{wilder2019melding,mandi2020interior}.
Other methods retain access to the original solver: for linear
objectives over fixed feasible sets, blackbox differentiation
constructs a piecewise-linear interpolation of the loss
~\citep{pogancic2020differentiation}, while random perturbations
smooth the optimizer's solution map
~\citep{berthet2020learning}.

An alternative is to replace the decision loss with a tractable
training objective. For linear objectives over fixed feasible
regions, SPO+ provides a convex surrogate whose subgradients
require optimization calls but not the optimizer's Jacobian
~\citep{elmachtoub2022smart}.
Contrastive and ranking losses compare objective values across
candidate feasible solutions, enabling differentiation when
these values are differentiable in the predicted parameters
~\citep{mulamba2020contrastive,mandi2020interior}.
Learned-loss approaches instead use solver evaluations to fit
differentiable approximations of decision loss, including
locally optimized decision losses
~\citep{shah2022decision} and landscape surrogates
~\citep{zharmagambetov2023landscape}.
These methods accommodate broader problem structures, but
train predictors through an approximation of the decision loss.

Motivated by stochastic smoothing, \citet{silvestri2026score}
use score function gradient estimation to train stochastic
predictors without differentiating through the optimizer
or requiring downstream convexity or linearity.

Our approach formulates decision-focused LLM post-training
as an RL problem, using negative decision regret as reward
without prescribing a particular policy optimization
algorithm. It targets expected regret directly, without
differentiating through the optimizer or requiring
downstream convexity or linearity.
Under our framework, the basic score-function estimator
of \citet{silvestri2026score} corresponds to REINFORCE
with negative decision loss as reward.

\subsection{LLMs as Predictors}
\label{sec: llms as predictors}

LLMs have demonstrated strong predictive capabilities across
tabular classification, numerical forecasting, and event
probability estimation. For tabular prediction, existing methods
represent features as natural-language descriptions
~\citep{hegselmann2023tabllm} or train language models on diverse
tables to support prediction on new datasets
~\citep{gardner2024large}. For time series forecasting, approaches
encode numerical sequences as text~\citep{gruver2023large}
or learn input representations for a frozen
LLM~\citep{jin2024time}. Beyond numerical inputs, language
models can extract predictive information from unstructured
text, with applications including clinical outcome prediction
from physician notes~\citep{jiang2023health}, stock-price
prediction from news headlines~\citep{lopez2026can},
and price distribution estimation from product
descriptions~\citep{vedula2025quantile}.
\citet{wang2024news} incorporate news analysis into time series
forecasting, while Time-MMD~\citep{liu2024time} and
Context is Key~\citep{williams2024context} provide benchmarks
for evaluating the use of textual context in forecasting.
Related approaches combine information retrieval, textual
reasoning, and tool use to estimate the probabilities of
real-world events~\citep{halawi2024approaching,hsieh2024reasoning}. Concurrent work by the \citet{percepta2026predictoptimize}
also studies the use of unstructured context in LLM-based
predict-then-optimize pipelines.

Our work builds on LLMs' ability to extract predictive information
from unstructured context, with a focus on predictions used as
parameters of downstream optimization problems. 

\subsection{RL for LLM Post-Training}
\label{sec: rl for LLM post-training}

RL is widely used to post-train language models using
feedback on generated responses, including human preferences
for instruction following
~\citep{ziegler2019fine,stiennon2020learning,ouyang2022training,touvron2023llama}
and verifiable feedback for reasoning tasks
~\citep{le2022coderl,guo2025deepseek}. Recent work develops policy optimization methods
for LLM post-training, exploring different approaches to
advantage estimation, policy updates, and computational
efficiency~\citep{schulman2017proximal,ahmadian2024back,shao2024deepseekmath}.
These developments provide multiple algorithmic choices
for learning from task-specific rewards.

Because obtaining reliable rewards can be costly in some tasks,
LLM post-training often relies on estimated feedback
~\citep{bai2022constitutional,lee2023rlaif}.
Errors in this feedback can distort the training objective,
motivating methods that improve reward learning from corrupted
preference labels~\citep{bukharin2024robust} or account for
reward-model uncertainty during policy
optimization~\citep{yan2024reward}.
A complementary approach corrects the observed rewards
under an explicit corruption model:
\citet{wang2020reinforcement} construct unbiased surrogate
rewards using a reward confusion matrix, while
\citet{mansouri2025noise} develop noise corrections under a binary reward-flip model.

Our work develops a framework for decision-focused
LLM post-training without prescribing a particular
RL algorithm.
To address expensive reward evaluation, we study
heuristic warm-starting and reward
correction.
Reward correction preserves the
expected policy gradient obtained with exact rewards,
even when the inexpensive cost estimates are biased.

\section{Decision-Focused Reinforcement Learning (DFRL)} \label{sec: dfrl}

\begin{figure*}[t]
    \centering
    \includegraphics[width=\textwidth]{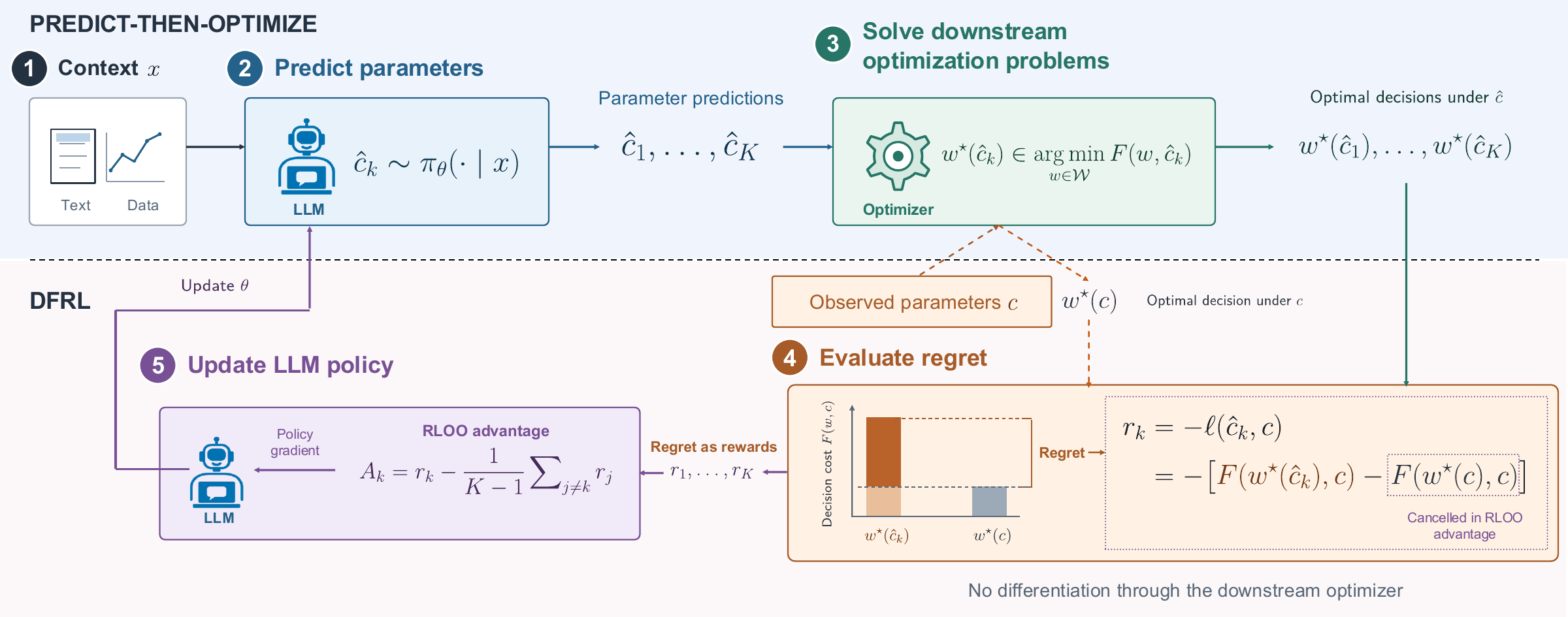}
    \caption{
    DFRL pipeline with RLOO as an example.
    \textbf{(1) Context.}
    The input $x$ provides task-relevant information,
    including text and structured data.
    \textbf{(2) Predict parameters.}
    The LLM independently samples $K$ parameter
    predictions $\hat c_1,\ldots,\hat c_K
    \sim\pi_\theta(\cdot\mid x)$.
    \textbf{(3) Solve downstream optimization problems.}
    Each prediction $\hat c_k$ is passed to the optimizer,
    which returns a decision $w^\star(\hat c_k)$ that is
    optimal under the predicted parameters.
    \textbf{(4) Evaluate regret.}
    Each decision is evaluated under the observed
    parameters $c$ and compared with the optimal decision
    under $c$, yielding reward $r_k=-\ell(\hat c_k,c)$.
    \textbf{(5) Update the LLM policy.}
    RLOO subtracts the average reward of the other
    $K-1$ predictions from $r_k$ to obtain advantage $A_k$.
    These advantages weight the policy gradients used
    to update $\theta$, without differentiating through
    the downstream optimizer.
    The hindsight cost $F(w^\star(c),c)$ is shared across
    predictions and cancels from the advantages, so
    solving under $c$ is unnecessary for the RLOO update.
    }
    \label{fig:dfrl_pipeline}
\end{figure*}

We now introduce DFRL, a post-training framework that enables LLMs to predict
optimization parameters using the negative regret of the
resulting decisions as reward. We illustrate the DFRL pipeline in
\Cref{fig:dfrl_pipeline}.

At each training iteration, we sample $(x,c)\sim\mathcal D$
and provide the context $x$ to the LLM, obtaining a parameter
prediction $\hat c\sim\pi_\theta(\cdot\mid x)$.
The downstream optimizer computes $w^\star(\hat c)$
according to~\eqref{eq:induced_decision}.
We evaluate this decision under the observed parameters $c$
and use its negative regret as reward:
\begin{equation*}
    r(\hat c,c):=-\ell(\hat c,c).
    \label{eq:dfrl_reward}
\end{equation*}
This reward is assigned to the LLM's generated response
and used to update the model through RL. The RL objective is
\begin{equation}
    \max_\theta
    J(\theta)
    :=
    \mathbb E_{(x,c)\sim\mathcal D}
    \left[
        \mathbb E_{\hat c\sim\pi_\theta(\cdot\mid x)}
        \big[r(\hat c,c)\big]
    \right]
    =
    -\mathcal L(\theta).
    \label{eq:dfrl_objective}
\end{equation}

Taking the gradient of~\eqref{eq:dfrl_objective} with respect
to $\theta$ and applying the score-function identity gives
\begin{equation}
    \nabla_\theta J(\theta)
    =
    \mathbb E_{(x,c)\sim\mathcal D}
    \left[
        \mathbb E_{\hat c\sim\pi_\theta(\cdot\mid x)}
        \left[
            r(\hat c,c)
            \nabla_\theta\log\pi_\theta(\hat c\mid x)
        \right]
    \right].
    \label{eq:dfrl_policy_gradient}
\end{equation}

The identity $J(\theta)=-\mathcal L(\theta)$ establishes
that maximizing expected reward minimizes expected decision
regret. Equation~\eqref{eq:dfrl_policy_gradient} further shows
that training requires only reward evaluations and
policy gradients, without differentiating through
the downstream optimizer. This formulation provides a common
objective for different RL post-training algorithms. \Cref{app:comparison of training objectives} compares the
expected-regret objective in Equation~\eqref{eq:dfrl_objective}
with those of existing decision-focused learning methods
and identifies conditions under which they coincide.

\subsection{Policy Optimization and Objective Alignment}
\label{sec:policy optimization and objective alignment}

Although DFRL does not prescribe a particular RL
algorithm, the choice of gradient estimator determines
whether its expected update matches
$\nabla_\theta J(\theta)$ in
Equation~\eqref{eq:dfrl_policy_gradient}.
We analyze RLOO~\citep{ahmadian2024back} as an instance
of DFRL and establish its alignment with the
expected-regret objective.

For each $(x,c)\sim\mathcal D$, we independently sample
$K\geq 2$ predictions
$\hat c_1,\ldots,\hat c_K\sim\pi_\theta(\cdot\mid x)$.
Let $r_k=-\ell(\hat c_k,c)$.
RLOO constructs the advantage
\begin{equation}
    A_k
    :=
    r_k-\frac{1}{K-1}\sum_{j\neq k}r_j.
    \label{eq:dfrl_rloo_advantage}
\end{equation}
Since all predictions share the same observed parameters
$c$, the hindsight cost $F(w^\star(c),c)$ cancels, giving
\begin{equation}
    A_k
    =
    -F(w^\star(\hat c_k),c)
    +
    \frac{1}{K-1}\sum_{j\neq k}
    F(w^\star(\hat c_j),c).
    \label{eq:dfrl_rloo_cost_advantage}
\end{equation}
Thus, the advantages can be computed directly from realized
decision costs without solving for $w^\star(c)$.

The resulting gradient estimator is
\begin{equation}
    \widehat G^{\mathrm{RLOO}}_\theta(x,c)
    :=
    \frac{1}{K}\sum_{k=1}^{K}
    A_k\,\nabla_\theta\log\pi_\theta(\hat c_k\mid x).
    \label{eq:dfrl_rloo}
\end{equation}
Conditional on $(x,c)$, the leave-one-out baseline
in~\eqref{eq:dfrl_rloo_advantage} is independent of
$\hat c_k$. Its contribution has zero expectation
by the zero-mean score identity, yielding
\begin{equation}
    \mathbb E[\widehat G^{\mathrm{RLOO}}_\theta(x,c)]
    =
    \nabla_\theta J(\theta)
    =
    -\nabla_\theta\mathcal L(\theta).
    \label{eq:dfrl_rloo_unbiased}
\end{equation}

Equation~\eqref{eq:dfrl_rloo_unbiased} establishes that
the RLOO estimator is unbiased for the negative gradient
of expected decision regret in Equation~\eqref{eq:dfrl_objective}.
Moreover, Equation~\eqref{eq:dfrl_rloo_cost_advantage}
allows its advantages to be computed without solving
the optimization problem under the observed parameters.
We give sufficient conditions for these updates to
approach stationarity of $\mathcal L(\theta)$ in
\Cref{app:rloo_convergence}.

\subsection{Efficient DFRL}
\label{sec:efficient DFRL}

Exact reward evaluation in DFRL requires solving a
downstream optimization problem for each sampled prediction.
A batch of $M$ instances with $K$ predictions per instance
therefore requires $MK$ optimization solves.
Even with parallel evaluation, each policy update must
wait for the batch's rewards, making optimization a
potential bottleneck in post-training.

Inexpensive reward estimates can enable more updates
within a fixed time budget, but using them directly
can bias the policy gradient relative to the
expected-regret objective in
Equation~\eqref{eq:dfrl_objective}.
We develop two approaches to this trade-off.
\emph{Heuristic warm-starting} uses approximate rewards
during an initial phase, accepting temporary gradient
bias in exchange for cheaper updates before switching
to exact-reward training.
\emph{Reward correction} combines approximate rewards
with randomized exact evaluations, preserving an
unbiased estimate of the expected-regret gradient
at the cost of additional variance.
The effectiveness of both approaches depends on
whether the computational savings compensate for
the introduced bias or variance.

For both approaches, consider a batch of independently
sampled instances $\{(x_m,c_m)\}_{m=1}^{M}$.
For each instance, we generate $K\geq 2$ independent
predictions $\hat c_{mk}\sim\pi_\theta(\cdot\mid x_m)$.
Let
\[
    C_{mk}:=F(w^\star(\hat c_{mk}),c_m)
\]
denote the exact realized decision cost, and let
$\widetilde C_{mk}$ denote an inexpensive estimate.
The following definitions describe both estimators;
exact costs need only be computed when required
by the training method.

Using Equation~\eqref{eq:dfrl_rloo_cost_advantage},
the exact and estimated advantages are
\begin{equation*}
    A_{mk}
    :=
    -C_{mk}
    +\frac{1}{K-1}\sum_{j\neq k}C_{mj},
    \qquad
    \widetilde A_{mk}
    :=
    -\widetilde C_{mk}
    +\frac{1}{K-1}\sum_{j\neq k}\widetilde C_{mj}.
    \label{eq:dfrl_estimated_advantage}
\end{equation*}
The leave-one-out baseline is computed within
each instance.

Write
$s_{mk}:=\nabla_\theta\log\pi_\theta(\hat c_{mk}\mid x_m)$.
The exact and inexpensive batch gradient estimators are
\begin{equation}
    G_\theta
    :=
    \frac{1}{MK}\sum_{m=1}^{M}\sum_{k=1}^{K}
    A_{mk}s_{mk},
    \qquad
    \widetilde G_\theta
    :=
    \frac{1}{MK}\sum_{m=1}^{M}\sum_{k=1}^{K}
    \widetilde A_{mk}s_{mk}.
    \label{eq:dfrl_estimated_gradient}
\end{equation}

\subsubsection{Heuristic Warm-Starting}
\label{sec:heuristic warm-starting}

For heuristic warm-starting, we construct the
inexpensive cost estimates using a
heuristic that returns a feasible decision
$\widetilde w(\hat c)\in W$:
\[
    \widetilde C_{mk}
    :=
    F(\widetilde w(\hat c_{mk}),c_m).
\]
During an initial training phase, we update the
policy using $\widetilde G_\theta$ from
Equation~\eqref{eq:dfrl_estimated_gradient},
without exact optimization solves.

These updates generally have nonzero bias relative
to the expected-regret gradient.
Define
\[
    b_\theta
    :=
    \mathbb E[\widetilde G_\theta]
    -\nabla_\theta J(\theta).
\]
Under the regularity assumptions in
\Cref{app:rloo_convergence}, the objective
$\mathcal L$ has an $L_J$-Lipschitz gradient.
Suppose that
$\|b_\theta\|\leq\varepsilon_{\mathrm h}$ and
$\mathbb E[
    \|\widetilde G_\theta
      -\mathbb E[\widetilde G_\theta]\|^2
]\leq\sigma_{\mathrm h}^2$
throughout the warm-start phase.
For a stochastic-gradient update
$\theta^+=\theta+\eta_{\mathrm h}\widetilde G_\theta$
with $0<\eta_{\mathrm h}\leq 1/L_J$, we show in
\Cref{app:heuristic_warm_starting} that
\begin{equation}
    \mathbb E[\mathcal L(\theta^+)\mid\theta]
    \leq
    \mathcal L(\theta)
    -
    \frac{\eta_{\mathrm h}}{2}
    \left(
        \|\nabla_\theta\mathcal L(\theta)\|^2
        -\varepsilon_{\mathrm h}^2
        -L_J\eta_{\mathrm h}\sigma_{\mathrm h}^2
    \right).
    \label{eq:heuristic_warm_descent}
\end{equation}
Thus, heuristic updates decrease the exact
expected-regret objective in expectation when the
squared gradient norm exceeds the combined bias
and sampling-noise terms.

This condition motivates using heuristic rewards
during an initial phase, when their bias may be
small relative to the expected-regret gradient
$\nabla_\theta\mathcal L(\theta)$.
As training progresses and this gradient
becomes smaller, the same bias can limit further
improvement.
We therefore switch to RLOO with exact rewards
(exact RLOO) after a warm-start period whose length is selected on validation data.

Heuristic warm-starting is beneficial when its
early progress compensates for the time taken
away from subsequent refinement.
\Cref{app:heuristic_warm_starting} provides the
descent analysis and a condition under which warm-starting improves the convergence guarantee of the subsequent exact-reward phase under a fixed time budget.

\subsubsection{Reward Correction}
\label{sec:reward correction theory}

Reward correction combines inexpensive cost estimates
with randomly selected exact evaluations to obtain
an unbiased estimator of the expected-regret gradient.
The inexpensive estimates $\widetilde C_{mk}$ may
come from heuristics or from
another approximation, such as a learned cost model.
For selected batches, we correct the estimated
advantages using their exact values; unselected
batches use inexpensive estimates alone.

For each batch, we draw a single
$I\sim\operatorname{Bernoulli}(p)$ with evaluation
probability $0<p\leq 1$.
If $I=1$, we solve the downstream problems for all
$MK$ predictions and compute their exact advantages;
if $I=0$, we skip exact evaluation for the whole batch.
The leave-one-out baselines are still computed within
each instance.
Define the corrected advantages as
\begin{equation}
    \widehat A_{mk}
    :=
    \begin{cases}
        \widetilde A_{mk}
        +\dfrac{A_{mk}-\widetilde A_{mk}}{p},
        & I=1,\\[6pt]
        \widetilde A_{mk},
        & I=0.
    \end{cases}
    \label{eq:dfrl_corrected_advantage}
\end{equation}
The resulting gradient estimator is
\begin{equation}
    \widehat G_\theta^{\mathrm{corr}}
    :=
    \frac{1}{MK}\sum_{m=1}^{M}\sum_{k=1}^{K}
    \widehat A_{mk}s_{mk}
    =
    \widetilde G_\theta
    +\frac{I}{p}(G_\theta-\widetilde G_\theta).
    \label{eq:dfrl_corrected_gradient}
\end{equation}

We retain the policy and regret-moment assumptions in
\Cref{app:rloo_convergence} and assume
\[
    \mathbb E\left[
        \frac{1}{MK}\sum_{m=1}^{M}\sum_{k=1}^{K}
        \widetilde A_{mk}^{\,2}
    \right]<\infty.
\]
For fixed $p>0$, these conditions ensure finite second
moments for the exact, inexpensive, and corrected
gradient estimators.

Conditional on the batch instances, predictions,
and cost estimates, expectation over $I$ gives
\begin{equation*}
    \mathbb E_I[\widehat A_{mk}]
    =
    \widetilde A_{mk}
    +p\,\frac{A_{mk}-\widetilde A_{mk}}{p}
    =
    A_{mk}.
    \label{eq:dfrl_advantage_unbiased}
\end{equation*}
Consequently,
\begin{equation*}
    \mathbb E_I[\widehat G_\theta^{\mathrm{corr}}]
    =
    G_\theta.
    \label{eq:dfrl_update_preservation}
\end{equation*}
Taking expectation over the batch and policy samples
then yields
\begin{equation}
    \mathbb E[\widehat G_\theta^{\mathrm{corr}}]
    =
    \mathbb E\!\left[
        \mathbb E_I[\widehat G_\theta^{\mathrm{corr}}]
    \right]
    =
    \mathbb E[G_\theta]
    =
    \nabla_\theta J(\theta)
    =
    -\nabla_\theta\mathcal L(\theta),
    \label{eq:dfrl_corrected_gradient_unbiased}
\end{equation}
where the unbiasedness of $G_\theta$ follows from
Equation~\eqref{eq:dfrl_rloo_unbiased}.
Thus, reward correction preserves the expected-regret
gradient without requiring unbiased inexpensive
cost estimates.

This unbiasedness does not guarantee faster training:
reducing $p$ saves exact evaluations but increases
the variance associated with discrepancies between
the exact and inexpensive gradient estimators.
In \Cref{app:correction_efficiency}, we characterize
when these computational savings improve gradient
estimation and convergence bounds under a fixed
expected time budget, accounting for the evaluation
probability, approximation quality, and evaluation cost.

\section{Experiments}
\label{sec:experiments}

We evaluate DFRL on synthetic optimization problems,
a computational-efficiency study, and an LLM-based
inventory planning task.

\subsection{Synthetic Experiments}
\label{sec:synthetic experiments}

\subsubsection{Experimental Setup}

\paragraph{Datasets.}
We consider three optimization problems: shortest path
on a $5\times5$ directed grid (LP), a 12-item knapsack
with two resource constraints (MILP), and quadratic
resource allocation with 12 decision variables (QP).
We generate contexts and optimization parameters using
the synthetic data generators in
PyEPO~\citep{tang2024}.
Each problem contains 1,000 training, 250 validation,
and 10,000 test instances.
Problem formulations and data-generation details are
provided in \Cref{app:synthetic_construction}.

\paragraph{Methods and evaluation.}
We compare DFRL using RLOO with predictors trained by
squared error, absolute error, and
SPO+~\citep{elmachtoub2022smart}.
All methods use affine predictors; DFRL uses their
outputs as the means of Gaussian prediction distributions.
We initialize DFRL from the squared-error predictor.
We report test regret using point predictions or,
for DFRL, predictive means.
More training details are
provided in \Cref{app:synthetic_training}.

\subsubsection{Main Results}

\begin{table}[htbp]
\centering
\caption{Test regret on synthetic problems. Lower is better.
DFRL results show the mean and sample standard
deviation across three training seeds.
DFRL starts from the squared-error
predictor.
Bold values indicate the lowest mean regret.
SPO+ is not applicable to quadratic allocation.}
\label{tab:synthetic_results}

\small
\setlength{\tabcolsep}{5pt}
\renewcommand{\arraystretch}{1.18}
\setlength{\aboverulesep}{0pt}
\setlength{\belowrulesep}{0pt}

\begin{tabular}{lccc}
\toprule
Method & Shortest path & Knapsack & Quadratic allocation \\
\midrule
Squared error
& 0.3173 & 1.4056 & 0.1814 \\
Absolute error
& 0.2834 & 1.4183 & 0.1552 \\
SPO+
& 0.2786 & 1.2489 & N/A \\
\midrule
\rowcolor{msftcard}
\textbf{DFRL (RLOO)}
& $\mathbf{0.2757} \pm 0.0018$
& $\mathbf{1.2178} \pm 0.0067$
& $\mathbf{0.1476} \pm 0.0005$ \\
\bottomrule
\end{tabular}
\end{table}

\paragraph{Decision quality.}
Table~\ref{tab:synthetic_results} shows that DFRL achieves
the lowest mean regret on all three problems, reducing regret
relative to the strongest baseline by 1.04\% on shortest path,
2.49\% on knapsack, and 4.90\% on quadratic allocation.
Relative to the squared-error predictor from which it is
initialized, the reductions are 13.1\%, 13.4\%, and 18.6\%.
DFRL also surpasses SPO+ on both linear problems without
using a decision-loss surrogate.

\subsection{Computational Efficiency Experiments}
\label{sec:computational_efficiency_experiments}

\subsubsection{Experimental Setup}

We evaluate computational efficiency on a larger
knapsack problem with 96 items and two resource
constraints, using 1,000 training, 250 validation,
and 10,000 test instances.
All DFRL methods use an affine Gaussian policy
initialized from the same least-squares predictor
and receive a one-hour training budget.
We compare exact RLOO, heuristic-only
training, heuristic warm-starting followed by
exact RLOO, and reward correction with
$p\in\{0.05,0.1,0.25,0.5,0.75,0.9\}$.
The methods share the optimizer settings and
the same cosine learning-rate schedule, and each
is evaluated over three training seeds shared across methods. Additional details appear in~\Cref{app:efficiency_experiments}.

\subsubsection{Main Results}

\begin{figure}[htbp]
\centering
\includegraphics[width=\linewidth]
{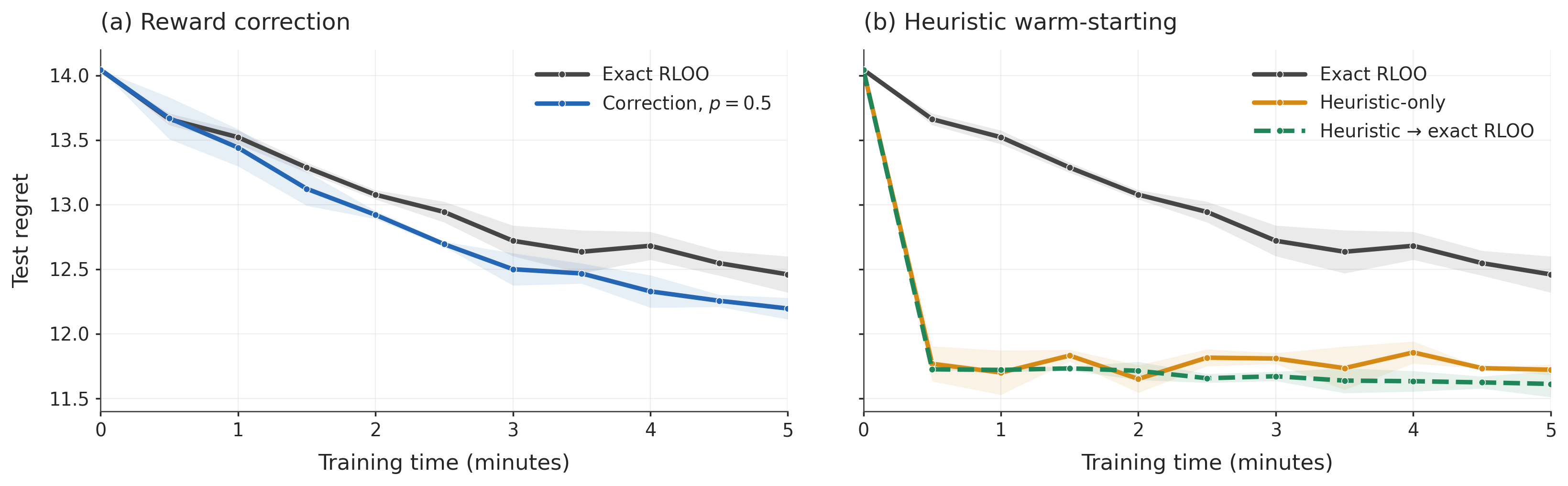}

\caption{Test regret comparison during the first five minutes of training on the 96-item knapsack problem.
(a) Reward correction with $p=0.5$ versus exact RLOO.
(b) Heuristic-only training and a heuristic
warm-start followed by exact RLOO.
Both panels use the same exact RLOO reference.
Curves show mean test regret over three training seeds shared
across methods; shading indicates $\pm$ one sample standard
deviation.}
\label{fig:efficiency_tradeoffs}
\end{figure}

Figure~\ref{fig:efficiency_tradeoffs} shows test regret
during the first five minutes of training, comparing
reward correction and heuristic warm-starting with
exact RLOO.
We also report one-hour training results, including
regret and update counts, in~\Cref{tab:efficiency_accounting}.

\paragraph{Reward correction.}
Figure~\ref{fig:efficiency_tradeoffs}(a) compares
correction with $p=0.5$ against exact RLOO.
Within five minutes, correction completes approximately
1,401 updates on average, compared with 741 for
exact RLOO, yielding about $1.9\times$ as many updates.
At this budget, correction achieves mean regret
$12.195 \pm 0.083$, compared with
$12.459 \pm 0.140$ for exact RLOO.
Thus, the cheaper updates translate into lower
five-minute mean regret.

\paragraph{Heuristic warm-starting.}
Figure~\ref{fig:efficiency_tradeoffs}(b) shows that
heuristic feedback produces rapid initial improvement.
Continued heuristic-only training subsequently
fluctuates, with limited further improvement over
the five-minute window.
A 10-second heuristic warm-start followed by
exact RLOO retains the early gains and
achieves regret $11.611 \pm 0.102$ at five minutes,
compared with $11.721 \pm 0.042$ for heuristic-only
training and $12.459 \pm 0.140$ for exact RLOO
from the same least-squares initialization.
Heuristic warm-starting speeds up early training, and exact rewards then improve on it.

\subsection{LLM-Based Inventory Planning}
\label{sec:llm_experiments}

\subsubsection{Experimental Setup}

\paragraph{Datasets.}
We construct an inventory planning dataset using transactions
and product descriptions from the H\&M Personalized Fashion
Recommendations dataset on Kaggle~\citep{h-and-m-personalized-fashion-recommendations}. 
Each instance contains 16 products, their descriptions and
attributes, and the preceding week's sales, with the following
week's sales as prediction targets. The dataset contains
4,013 training, 527 validation, and 1,251 test instances.

For each instance, the optimizer chooses an order quantity for
every product from a discrete menu, subject to a shared purchasing
budget, to minimize overage and shortage costs under the predicted
sales. 
We evaluate decisions
by regret under observed sales.
More details are
in \Cref{app:inventory_construction}.

\paragraph{Baselines.}
We compare against linear regression and XGBoost, each
trained with product descriptions.
Both models use historical sales, product attributes,
calendar variables, and the instance’s cost coefficients, order menus, and budget.
Descriptions are encoded using TF--IDF
features~\citep{salton1988term}, with
dimensionality reduction for XGBoost.
We also perform SFT on Qwen3.5-4B
using observed sales as target
responses. 
\Cref{app:sft_prediction_loss} explains how SFT constitutes
prediction-loss training for LLMs.
Hyperparameters are detailed in
\Cref{app:inventory_training}.

\paragraph{DFRL.}
We initialize Qwen3.5-4B from the SFT checkpoint and
post-train it using RLOO or GRPO.
For each training instance, the LLM samples eight sales
predictions using the prompt in
\Cref{app:inventory_prompts}.
Each prediction is passed to the inventory optimizer,
and the resulting decision's negative regret serves
as reward. We also evaluate negative SPO+ as a
surrogate-reward ablation.

In the thinking-enabled variant, the LLM writes a reasoning trace
before its sales predictions. We first perform reasoning-trace SFT:
GPT-6-luna receives each training context together with its observed
sales and writes an explanation of how the context supports them
(\Cref{app:reasoning_teacher_prompt}). The LLM is then trained with
SFT to produce this explanation followed by the sales, without
seeing the observed sales in its input. We then continue with RLOO,
with thinking enabled.
Further training details are provided in
\Cref{app:inventory_training}.

\subsubsection{Main Results}

\begin{table}[htbp]
\centering
\caption{Inventory planning results on 1,251 test instances.
All methods include product descriptions.
The Objective column gives the training loss for the baselines
and the reward for DFRL; $-$SPO+ denotes the negative SPO+ loss
used as a surrogate reward.
For the linear model, SPO+ training starts from the squared-error
predictor and validation selects that initial checkpoint, so the
two are reported in one row.
``Thinking'' denotes reasoning-trace SFT followed by
RL with thinking enabled.
Regret measures decision quality and MAE measures sales
prediction accuracy; lower is better.
Bold values indicate the best result in each metric.}
\label{tab:llm_inventory_results}

\small
\setlength{\tabcolsep}{5pt}
\renewcommand{\arraystretch}{1.15}
\setlength{\aboverulesep}{0pt}
\setlength{\belowrulesep}{0pt}

\begin{tabular}{llrr}
\toprule
Method & Objective & Regret$\downarrow$ & MAE$\downarrow$ \\
\midrule
Linear & Squared error / SPO+ & 204.96 & 11.609 \\
XGBoost & Squared error & 183.22 & 11.410 \\
\midrule
\rowcolor{promptbg}
SFT & Cross-entropy & 234.01 & 11.726 \\
\midrule
DFRL (RLOO) & $-\mathrm{SPO+}$ & 222.11 & 11.316 \\
DFRL (GRPO) & $-\mathrm{SPO+}$ & 200.58 & 11.095 \\
\midrule
\rowcolor{msftcard}
\textbf{DFRL (GRPO)}
& $-\mathrm{Regret}$ & 174.67 & 11.470 \\
\rowcolor{msftcard}
\textbf{DFRL (RLOO)}
& $-\mathrm{Regret}$ & 172.72 & 11.395 \\
\rowcolor{msftcard}
\textbf{DFRL (RLOO, thinking)}
& $-\mathrm{Regret}$ & \textbf{160.53} & \textbf{10.950} \\
\bottomrule
\end{tabular}
\end{table}

\paragraph{Effectiveness of DFRL.}
With negative regret rewards, DFRL using RLOO reduces
mean regret from 234.01 to 172.72, improving on SFT by
26.2\% and on the strongest conventional forecasting
baseline (XGBoost) by 5.7\%.
The thinking-enabled variant achieves the lowest regret
and MAE, with regret reductions of 31.4\% relative to
SFT and 12.4\% relative to the strongest conventional
baseline.
Negative regret rewards also outperform negative SPO+
rewards under both RLOO and GRPO, supporting the use
of decision regret directly as training feedback.

\paragraph{Prediction accuracy and decision quality.}
Lower prediction error does not consistently imply
lower regret.
For example, under RLOO, switching from negative SPO+
to negative regret increases MAE from 11.316 to 11.395
while reducing regret from 222.11 to 172.72.
The same pattern occurs under GRPO.

\section{Conclusion and Discussion}
\label{sec:conclusion and discussion}

We propose decision-focused reinforcement learning
(DFRL), a framework for post-training LLMs to predict
optimization parameters using negative downstream
decision regret as reward, without differentiating
through the downstream optimizer.
Experiments show improved decision quality on
three synthetic optimization problems and an
LLM-based inventory planning task.
Our best inventory configuration combines
reasoning-trace SFT with thinking-enabled DFRL
and reduces regret by 31.4\% relative to the
SFT baseline and 12.4\% relative to the strongest
evaluated conventional forecasting baseline.

To address expensive reward evaluation, we
develop two approaches.
Heuristic warm-starting uses inexpensive,
potentially biased feedback to accelerate
initial learning before switching to exact-reward
refinement.
Reward correction preserves the expected policy
gradient while trading additional estimator
variance for cheaper updates.
Our analysis gives sufficient conditions under
which these computational savings improve
convergence bounds.
In the 96-item knapsack experiment, heuristic
warm-starting accelerates early improvement.
Reward correction also improves regret at
selected short training budgets.

DFRL applies wherever unstructured data informs an optimization problem, such as supply chain planning, energy scheduling, and resource allocation. Heuristic warm-starting and reward correction can also be useful in other RL settings where exact rewards are expensive but cheap approximations exist.

\section*{Acknowledgement}

We thank Thinking Machines Lab for  supporting
this research through the Tinker Research Grant.

\bibliographystyle{assets/plainnat}
\bibliography{citation}

@article{sadana2025survey,
  title={A survey of contextual optimization methods for decision-making under uncertainty},
  author={Sadana, Utsav and Chenreddy, Abhilash and Delage, Erick and Forel, Alexandre and Frejinger, Emma and Vidal, Thibaut},
  journal={European Journal of Operational Research},
  volume={320},
  number={2},
  pages={271--289},
  year={2025},
  publisher={Elsevier}
}

@article{donti2017task,
  title={Task-based end-to-end model learning in stochastic optimization},
  author={Donti, Priya and Amos, Brandon and Kolter, J Zico},
  journal={Advances in neural information processing systems},
  volume={30},
  year={2017}
}

@article{elmachtoub2022smart,
  title={Smart “predict, then optimize”},
  author={Elmachtoub, Adam N and Grigas, Paul},
  journal={Management science},
  volume={68},
  number={1},
  pages={9--26},
  year={2022},
  publisher={INFORMS}
}

@inproceedings{wilder2019melding,
  title={Melding the data-decisions pipeline: Decision-focused learning for combinatorial optimization},
  author={Wilder, Bryan and Dilkina, Bistra and Tambe, Milind},
  booktitle={Proceedings of the AAAI conference on artificial intelligence},
  volume={33},
  number={01},
  pages={1658--1665},
  year={2019}
}

@article{silver2016mastering,
  title={Mastering the game of Go with deep neural networks and tree search},
  author={Silver, David and Huang, Aja and Maddison, Chris J and Guez, Arthur and Sifre, Laurent and Van Den Driessche, George and Schrittwieser, Julian and Antonoglou, Ioannis and Panneershelvam, Veda and Lanctot, Marc and others},
  journal={nature},
  volume={529},
  number={7587},
  pages={484--489},
  year={2016},
  publisher={Nature Publishing Group UK, London}
}

@article{bello2016neural,
  title={Neural combinatorial optimization with reinforcement learning},
  author={Bello, Irwan and Pham, Hieu and Le, Quoc V and Norouzi, Mohammad and Bengio, Samy},
  journal={arXiv preprint arXiv:1611.09940},
  year={2016}
}

@article{guo2025deepseek,
  title={Deepseek-r1: Incentivizing reasoning capability in llms via reinforcement learning},
  author={Guo, Daya and Yang, Dejian and Zhang, Haowei and Song, Junxiao and Wang, Peiyi and Zhu, Qihao and Xu, Runxin and Zhang, Ruoyu and Ma, Shirong and Bi, Xiao and others},
  journal={arXiv preprint arXiv:2501.12948},
  year={2025}
}

@article{williams1992simple,
  title={Simple statistical gradient-following algorithms for connectionist reinforcement learning},
  author={Williams, Ronald J},
  journal={Machine learning},
  volume={8},
  number={3},
  pages={229--256},
  year={1992},
  publisher={Springer}
}

@article{gruver2023large,
  title={Large language models are zero-shot time series forecasters},
  author={Gruver, Nate and Finzi, Marc and Qiu, Shikai and Wilson, Andrew G},
  journal={Advances in neural information processing systems},
  volume={36},
  pages={19622--19635},
  year={2023}
}

@article{wang2024news,
  title={From news to forecast: Integrating event analysis in llm-based time series forecasting with reflection},
  author={Wang, Xinlei and Feng, Maike and Qiu, Jing and Gu, Jinjin and Zhao, Junhua},
  journal={Advances in Neural Information Processing Systems},
  volume={37},
  pages={58118--58153},
  year={2024}
}

@article{gardner2024large,
  title={Large scale transfer learning for tabular data via language modeling},
  author={Gardner, Josh and Perdomo, Juan C and Schmidt, Ludwig},
  journal={Advances in Neural Information Processing Systems},
  volume={37},
  pages={45155--45205},
  year={2024}
}

@inproceedings{hegselmann2023tabllm,
  title={Tabllm: Few-shot classification of tabular data with large language models},
  author={Hegselmann, Stefan and Buendia, Alejandro and Lang, Hunter and Agrawal, Monica and Jiang, Xiaoyi and Sontag, David},
  booktitle={International conference on artificial intelligence and statistics},
  pages={5549--5581},
  year={2023},
  organization={PMLR}
}

@inproceedings{jin2024time,
  title={Time-llm: Time series forecasting by reprogramming large language models},
  author={Jin, Ming and Wang, Shiyu and Ma, Lintao and Chu, Zhixuan and Zhang, James and Shi, Xiaoming and Chen, Pin-Yu and Liang, Yuxuan and Li, Yuan-Fang and Pan, Shirui and others},
  booktitle={International conference on learning representations},
  volume={2024},
  pages={23857--23880},
  year={2024}
}

@article{halawi2024approaching,
  title={Approaching human-level forecasting with language models},
  author={Halawi, Danny and Zhang, Fred and Yueh-Han, Chen and Steinhardt, Jacob},
  journal={Advances in Neural Information Processing Systems},
  volume={37},
  pages={50426--50468},
  year={2024}
}

@article{ouyang2022training,
  title={Training language models to follow instructions with human feedback},
  author={Ouyang, Long and Wu, Jeffrey and Jiang, Xu and Almeida, Diogo and Wainwright, Carroll and Mishkin, Pamela and Zhang, Chong and Agarwal, Sandhini and Slama, Katarina and Ray, Alex and others},
  journal={Advances in neural information processing systems},
  volume={35},
  pages={27730--27744},
  year={2022}
}

@article{touvron2023llama,
  title={Llama 2: Open foundation and fine-tuned chat models},
  author={Touvron, Hugo and Martin, Louis and Stone, Kevin and Albert, Peter and Almahairi, Amjad and Babaei, Yasmine and Bashlykov, Nikolay and Batra, Soumya and Bhargava, Prajjwal and Bhosale, Shruti and others},
  journal={arXiv preprint arXiv:2307.09288},
  year={2023}
}

@article{grattafiori2024llama,
  title={The llama 3 herd of models},
  author={Grattafiori, Aaron and Dubey, Abhimanyu and Jauhri, Abhinav and Pandey, Abhinav and Kadian, Abhishek and Al-Dahle, Ahmad and Letman, Aiesha and Mathur, Akhil and Schelten, Alan and Vaughan, Alex and others},
  journal={arXiv preprint arXiv:2407.21783},
  year={2024}
}

@article{silvestri2026score,
  title={Score function gradient estimation to widen the applicability of decision-focused learning},
  author={Silvestri, Mattia and Berden, Senne and Signorelli, Gaetano and Mahmuto{\u{g}}ullar{\i}, Ali {\.I}rfan and Mandi, Jayanta and Amos, Brandon and Guns, Tias and Lombardi, Michele},
  journal={Journal of Artificial Intelligence Research},
  volume={85},
  year={2026}
}

@inproceedings{amos2017optnet,
  title={Optnet: Differentiable optimization as a layer in neural networks},
  author={Amos, Brandon and Kolter, J Zico},
  booktitle={International conference on machine learning},
  pages={136--145},
  year={2017},
  organization={PMLR}
}

@article{agrawal2019differentiable,
  title={Differentiable convex optimization layers},
  author={Agrawal, Akshay and Amos, Brandon and Barratt, Shane and Boyd, Stephen and Diamond, Steven and Kolter, J Zico},
  journal={Advances in neural information processing systems},
  volume={32},
  year={2019}
}

@article{mandi2020interior,
  title={Interior point solving for lp-based prediction+ optimisation},
  author={Mandi, Jayanta and Guns, Tias},
  journal={Advances in Neural Information Processing Systems},
  volume={33},
  pages={7272--7282},
  year={2020}
}

@inproceedings{pogancic2020differentiation,
  title={Differentiation of Blackbox Combinatorial Solvers.},
  author={Pogancic, Marin Vlastelica and Paulus, Anselm and Musil, Vit and Martius, Georg and Rolinek, Michal},
  booktitle={ICLR},
  year={2020}
}

@article{berthet2020learning,
  title={Learning with differentiable pertubed optimizers},
  author={Berthet, Quentin and Blondel, Mathieu and Teboul, Olivier and Cuturi, Marco and Vert, Jean-Philippe and Bach, Francis},
  journal={Advances in neural information processing systems},
  volume={33},
  pages={9508--9519},
  year={2020}
}

@article{mulamba2020contrastive,
  title={Contrastive losses and solution caching for predict-and-optimize},
  author={Mulamba, Maxime and Mandi, Jayanta and Diligenti, Michelangelo and Lombardi, Michele and Bucarey, Victor and Guns, Tias},
  journal={arXiv preprint arXiv:2011.05354},
  year={2020}
}

@article{shah2022decision,
  title={Decision-focused learning without decision-making: Learning locally optimized decision losses},
  author={Shah, Sanket and Wang, Kai and Wilder, Bryan and Perrault, Andrew and Tambe, Milind},
  journal={Advances in Neural Information Processing Systems},
  volume={35},
  pages={1320--1332},
  year={2022}
}

@article{zharmagambetov2023landscape,
  title={Landscape surrogate: Learning decision losses for mathematical optimization under partial information},
  author={Zharmagambetov, Arman and Amos, Brandon and Ferber, Aaron and Huang, Taoan and Dilkina, Bistra and Tian, Yuandong},
  journal={Advances in Neural Information Processing Systems},
  volume={36},
  pages={27332--27350},
  year={2023}
}

@article{jiang2023health,
  title={Health system-scale language models are all-purpose prediction engines},
  author={Jiang, Lavender Yao and Liu, Xujin Chris and Nejatian, Nima Pour and Nasir-Moin, Mustafa and Wang, Duo and Abidin, Anas and Eaton, Kevin and Riina, Howard Antony and Laufer, Ilya and Punjabi, Paawan and others},
  journal={Nature},
  volume={619},
  number={7969},
  pages={357--362},
  year={2023},
  publisher={Nature Publishing Group UK, London}
}

@inproceedings{vedula2025quantile,
  title={Quantile regression with large language models for price prediction},
  author={Vedula, Nikhita and Dhyani, Dushyanta and Jalali, Laleh and Oreshkin, Boris N and Bayati, Mohsen and Malmasi, Shervin},
  booktitle={Findings of the Association for Computational Linguistics: ACL 2025},
  pages={12396--12415},
  year={2025}
}

@article{lopez2026can,
  title={Can chatgpt forecast stock price movements? return predictability and large language models},
  author={Lopez-Lira, Alejandro and Tang, Yuehua},
  journal={Journal of Financial Economics},
  volume={184},
  pages={104335},
  year={2026},
  publisher={Elsevier}
}

@article{liu2024time,
  title={Time-mmd: Multi-domain multimodal dataset for time series analysis},
  author={Liu, Haoxin and Xu, Shangqing and Zhao, Zhiyuan and Kong, Lingkai and Kamarthi, Harshavardhan and Sasanur, Aditya B and Sharma, Megha and Cui, Jiaming and Wen, Qingsong and Zhang, Chao and others},
  journal={Advances in Neural Information Processing Systems},
  volume={37},
  pages={77888--77933},
  year={2024}
}

@article{williams2024context,
  title={Context is key: A benchmark for forecasting with essential textual information},
  author={Williams, Andrew Robert and Ashok, Arjun and Marcotte, {\'E}tienne and Zantedeschi, Valentina and Subramanian, Jithendaraa and Riachi, Roland and Requeima, James and Lacoste, Alexandre and Rish, Irina and Chapados, Nicolas and others},
  journal={arXiv preprint arXiv:2410.18959},
  year={2024}
}

@inproceedings{hsieh2024reasoning,
  title={Reasoning and tools for human-level forecasting},
  author={Hsieh, Elvis and Fu, Preston and Chen, Jonathan},
  booktitle={Proceedings of the Workshop on the Future of Event Detection (FuturED)},
  pages={48--57},
  year={2024}
}

@article{ziegler2019fine,
  title={Fine-tuning language models from human preferences},
  author={Ziegler, Daniel M and Stiennon, Nisan and Wu, Jeffrey and Brown, Tom B and Radford, Alec and Amodei, Dario and Christiano, Paul and Irving, Geoffrey},
  journal={arXiv preprint arXiv:1909.08593},
  year={2019}
}

@article{stiennon2020learning,
  title={Learning to summarize with human feedback},
  author={Stiennon, Nisan and Ouyang, Long and Wu, Jeffrey and Ziegler, Daniel and Lowe, Ryan and Voss, Chelsea and Radford, Alec and Amodei, Dario and Christiano, Paul F},
  journal={Advances in neural information processing systems},
  volume={33},
  pages={3008--3021},
  year={2020}
}

@article{le2022coderl,
  title={Coderl: Mastering code generation through pretrained models and deep reinforcement learning},
  author={Le, Hung and Wang, Yue and Gotmare, Akhilesh Deepak and Savarese, Silvio and Hoi, Steven Chu Hong},
  journal={Advances in Neural Information Processing Systems},
  volume={35},
  pages={21314--21328},
  year={2022}
}

@article{schulman2017proximal,
  title={Proximal policy optimization algorithms},
  author={Schulman, John and Wolski, Filip and Dhariwal, Prafulla and Radford, Alec and Klimov, Oleg},
  journal={arXiv preprint arXiv:1707.06347},
  year={2017}
}

@inproceedings{ahmadian2024back,
  title={Back to basics: Revisiting REINFORCE-style optimization for learning from human feedback in LLMs},
  author={Ahmadian, Arash and Cremer, Chris and Gall{\'e}, Matthias and Fadaee, Marzieh and Kreutzer, Julia and Pietquin, Olivier and {\"U}st{\"u}n, Ahmet and Hooker, Sara},
  booktitle={Proceedings of the 62nd Annual Meeting of the Association for Computational Linguistics (Volume 1: Long Papers)},
  pages={12248--12267},
  year={2024}
}

@article{shao2024deepseekmath,
  title={Deepseekmath: Pushing the limits of mathematical reasoning in open language models},
  author={Shao, Zhihong and Wang, Peiyi and Zhu, Qihao and Xu, Runxin and Song, Junxiao and Bi, Xiao and Zhang, Haowei and Zhang, Mingchuan and Li, YK and Wu, Yang and others},
  journal={arXiv preprint arXiv:2402.03300},
  year={2024}
}

@article{bai2022constitutional,
  title={Constitutional ai: Harmlessness from ai feedback},
  author={Bai, Yuntao and Kadavath, Saurav and Kundu, Sandipan and Askell, Amanda and Kernion, Jackson and Jones, Andy and Chen, Anna and Goldie, Anna and Mirhoseini, Azalia and McKinnon, Cameron and others},
  journal={arXiv preprint arXiv:2212.08073},
  year={2022}
}

@article{lee2023rlaif,
  title={Rlaif vs. rlhf: Scaling reinforcement learning from human feedback with ai feedback},
  author={Lee, Harrison and Phatale, Samrat and Mansoor, Hassan and Mesnard, Thomas and Ferret, Johan and Lu, Kellie and Bishop, Colton and Hall, Ethan and Carbune, Victor and Rastogi, Abhinav and others},
  journal={arXiv preprint arXiv:2309.00267},
  year={2023}
}

@article{bukharin2024robust,
  title={Robust reinforcement learning from corrupted human feedback},
  author={Bukharin, Alexander and Hong, Ilgee and Jiang, Haoming and Li, Zichong and Zhang, Qingru and Zhang, Zixuan and Zhao, Tuo},
  journal={Advances in Neural Information Processing Systems},
  volume={37},
  pages={124093--124113},
  year={2024}
}

@article{yan2024reward,
  title={Reward-robust rlhf in llms},
  author={Yan, Yuzi and Lou, Xingzhou and Li, Jialian and Zhang, Yiping and Xie, Jian and Yu, Chao and Wang, Yu and Yan, Dong and Shen, Yuan},
  journal={arXiv preprint arXiv:2409.15360},
  year={2024}
}

@inproceedings{wang2020reinforcement,
  title={Reinforcement learning with perturbed rewards},
  author={Wang, Jingkang and Liu, Yang and Li, Bo},
  booktitle={Proceedings of the AAAI conference on artificial intelligence},
  volume={34},
  number={04},
  pages={6202--6209},
  year={2020}
}

@article{mansouri2025noise,
  title={Noise-corrected grpo: From noisy rewards to unbiased gradients},
  author={Mansouri, Omar El and Izzati, Fathinah Asma and Seddik, Mohamed El Amine and Lahlou, Salem},
  journal={arXiv preprint arXiv:2510.18924},
  year={2025}
}

@article{ghadimi2013stochastic,
  title={Stochastic first-and zeroth-order methods for nonconvex stochastic programming},
  author={Ghadimi, Saeed and Lan, Guanghui},
  journal={SIAM journal on optimization},
  volume={23},
  number={4},
  pages={2341--2368},
  year={2013},
  publisher={SIAM}
}

@misc{h-and-m-personalized-fashion-recommendations,
    author = {Carlos García Ling and ElizabethHMGroup and FridaRim and inversion and Jaime Ferrando and Maggie and neuraloverflow and xlsrln},
    title = {H\&M Personalized Fashion Recommendations},
    year = {2022},
    howpublished = {\url{https://www.kaggle.com/competitions/h-and-m-personalized-fashion-recommendations}},
    note = {Kaggle}
}

@article{salton1988term,
  title={Term-weighting approaches in automatic text retrieval},
  author={Salton, Gerard and Buckley, Christopher},
  journal={Information processing \& management},
  volume={24},
  number={5},
  pages={513--523},
  year={1988},
  publisher={Elsevier}
}

@misc{percepta2026predictoptimize,
  author = {{Percepta AI Research Team}},
  title  = {{LLM} Predict-then-Optimize: Extracting Maximum Information from Unstructured Contexts},
  year   = {2026},
  note   = {Working paper}
}

@article{tang2024,
  title={PyEPO: a PyTorch-based end-to-end predict-then-optimize library for linear and integer programming},
  author={Tang, Bo and Khalil, Elias B},
  journal={Mathematical Programming Computation},
  issn={1867-2957},
  doi={10.1007/s12532-024-00255-x},
  year={2024},
  month={July},
  publisher={Springer}
}

\clearpage
\newpage
\appendix
\appendix
\crefalias{section}{appendix}

\section{Theoretical Analysis}
\label[appendix]{app:theoretical analysis}

\subsection{Comparison of Training Objectives}
\label[appendix]{app:comparison of training objectives}

Decision-focused learning methods share the goal of training
predictive models according to the quality of their induced
decisions. They differ in how predictions are represented
and used in the downstream optimization problem.
We compare these formulations using the feasible set
$\mathcal W$, objective function $F(w,c)$, and data
distribution $\mathcal D$ defined in
Section~\ref{sec:Background}.

\paragraph{A common decision-focused objective.}
Let $w_\theta(x,\xi)\in\mathcal W$ denote the decision
induced by a predictive model and its downstream optimization
procedure, where $\xi$ represents any sampling randomness
and is independent of $(x,c)$.
The common target is to minimize expected decision regret:
\begin{equation}
    \min_\theta
    \mathbb E_{(x,c)\sim\mathcal D,\,\xi}
    \left[
        F(w_\theta(x,\xi),c)
        -
        F(w^\star(c),c)
    \right].
    \label{eq:app_common_objective}
\end{equation}
For a deterministic decision, $w_\theta$ does not depend
on $\xi$.
Since the hindsight cost $F(w^\star(c),c)$ is independent
of $\theta$, minimizing expected regret is equivalent
to minimizing expected realized cost.
The formulations below specify different ways to
construct $w_\theta$.

\paragraph{Predicting a distribution for stochastic optimization.}
In task-based learning for stochastic
optimization~\citep{donti2017task}, a model predicts a
distribution $q_\theta(\cdot\mid x)$ and selects a decision
by minimizing expected cost under that distribution.
Specializing this approach to our fixed feasible set gives
\begin{equation*}
    w_{\mathrm{dist},\theta}(x)
    \in
    \arg\min_{w\in\mathcal W}
    \mathbb E_{\tilde c\sim q_\theta(\cdot\mid x)}
    [F(w,\tilde c)].
    \label{eq:app_distribution_decision}
\end{equation*}
Its expected-regret objective is
\begin{equation*}
    \mathcal L_{\mathrm{dist}}(\theta)
    :=
    \mathbb E_{(x,c)\sim\mathcal D}
    \left[
        F(w_{\mathrm{dist},\theta}(x),c)
        -
        F(w^\star(c),c)
    \right].
    \label{eq:app_distribution_objective}
\end{equation*}
Although the prediction is a distribution, the optimization
procedure selects one decision for each context $x$,
using a fixed rule to resolve ties.
The expectation over predicted parameters is therefore
part of the decision construction.

\paragraph{Predicting a deterministic parameter vector.}
In decision-focused learning using point
predictions~\citep{elmachtoub2022smart,wilder2019melding}, a point predictor produces $\hat c=f_\theta(x)$, and
the downstream optimizer selects $w^\star(f_\theta(x))$.
The resulting objective is
\begin{equation*}
    \mathcal L_{\mathrm{point}}(\theta)
    :=
    \mathbb E_{(x,c)\sim\mathcal D}
    \left[\ell(f_\theta(x),c)\right].
    \label{eq:app_point_objective}
\end{equation*}
For $F(w,c)=c^\top w$, this is the expected SPO
loss~\citep{elmachtoub2022smart}.
SPO+ provides a convex surrogate for training,
so its surrogate objective must be distinguished
from the target regret objective above.

\paragraph{Sampling parameter predictions.}
In DFRL, the model samples
$\hat c\sim\pi_\theta(\cdot\mid x)$, and the downstream
optimizer selects $w^\star(\hat c)$ for that prediction.
The induced decision is therefore stochastic, and
the common objective becomes
\begin{equation*}
    \mathcal L(\theta)
    =
    \mathbb E_{(x,c)\sim\mathcal D}
    \left[
        \mathbb E_{\hat c\sim\pi_\theta(\cdot\mid x)}
        [\ell(\hat c,c)]
    \right].
    \label{eq:app_stochastic_objective}
\end{equation*}
DFRL optimizes this objective by using negative regret
as the RL reward.
Here, the expectation averages regret over decisions
induced by sampled predictions; it is not taken inside
the downstream optimization problem.
Deterministic parameter prediction is included by setting
$\pi_\theta(\cdot\mid x)=\delta_{f_\theta(x)}$.

\paragraph{Relationship between the formulations.}
These formulations share the decision-focused criterion
in~\eqref{eq:app_common_objective}, but generally induce
different decisions and therefore different objective
functions of $\theta$.
Their relationship depends on the downstream problem
and the predictive models.
We next give sufficient conditions under which their
induced decisions
coincide.

Fix $\theta$, assume that the relevant expectations are finite,
and assume unique optimal solutions for the comparisons below.
Each condition is required to hold for
$\mathcal D_X$-almost every context $x$, where $\mathcal D_X$
is the marginal distribution of $x$ under $\mathcal D$.

\begin{enumerate}
    \item \textbf{Distribution-based vs.\ point prediction.}
    If $F(w,c)=c^\top w$ and
    $f_\theta(x)=
    \mathbb E_{\tilde c\sim q_\theta(\cdot\mid x)}[\tilde c]$,
    then, for every feasible decision $w$,
    \[
        \mathbb E_{\tilde c\sim q_\theta(\cdot\mid x)}
        [F(w,\tilde c)]
        =
        f_\theta(x)^\top w.
    \]
    Optimizing under the predicted distribution is therefore
    the same optimization problem as optimizing under its
    mean. Consequently,
    \[
        w_{\mathrm{dist},\theta}(x)
        =
        w^\star(f_\theta(x)),
        \qquad
        \mathcal L_{\mathrm{dist}}(\theta)
        =
        \mathcal L_{\mathrm{point}}(\theta).
    \]

    \item \textbf{Point vs.\ stochastic prediction.}
    Suppose
    \[
        \Pr_{\hat c\sim\pi_\theta(\cdot\mid x)}
        \!\left(
            w^\star(\hat c)=w^\star(f_\theta(x))
        \right)=1.
    \]
    Every sampled prediction then produces the same decision
    as the point prediction. Hence,
    \[
        \mathcal L(\theta)=\mathcal L_{\mathrm{point}}(\theta).
    \]
    This holds when
    $\pi_\theta(\cdot\mid x)=\delta_{f_\theta(x)}$,
    but also when different sampled predictions lead to the
    same decision.

    \item \textbf{Distribution-based vs.\ stochastic prediction.}
    Suppose $q_\theta(\cdot\mid x)=\pi_\theta(\cdot\mid x)$
    and there exists $w_0(x)\in\mathcal W$ such that
    \[
        \Pr_{\hat c\sim\pi_\theta(\cdot\mid x)}
        \!\left(w^\star(\hat c)=w_0(x)\right)=1.
    \]
    Since $w_0(x)$ minimizes cost for almost every sampled
    parameter vector, it also minimizes their expected cost:
    \[
        \mathbb E_{\hat c\sim\pi_\theta(\cdot\mid x)}
        [F(w_0(x),\hat c)]
        \leq
        \mathbb E_{\hat c\sim\pi_\theta(\cdot\mid x)}
        [F(w,\hat c)]
        \quad\text{for all }w\in\mathcal W.
    \]
    Thus, both formulations select $w_0(x)$, and
    \[
        \mathcal L_{\mathrm{dist}}(\theta)=\mathcal L(\theta).
    \]
\end{enumerate}

\subsection{Convergence under Decision-Regret Rewards}
\label[appendix]{app:rloo_convergence}

We verify sufficient conditions for applying standard
nonconvex stochastic-gradient convergence
results~\citep{ghadimi2013stochastic} to DFRL.
The analysis concerns the RLOO estimator in
Equation~\eqref{eq:dfrl_rloo} and the expected-regret
objective $\mathcal L(\theta)$.

\paragraph{LLM sampling and reward evaluation.}
For this analysis, we make the response-level sampling
explicit.
Let $p_\theta(y\mid x)$ denote the probability of an LLM
response $y$, and let $g$ be the fixed parsing function
that produces $\hat c=g(y)$.
The induced distribution of $\hat c$ is
$\pi_\theta(\cdot\mid x)$.
The response receives reward
$r(y,c)=-\ell(g(y),c)$.
In the LLM implementation, the score in
Equation~\eqref{eq:dfrl_rloo} is evaluated as
$\nabla_\theta\log p_\theta(y\mid x)$.
Applying the score-function identity at the response level
gives the same expected gradient
$\nabla_\theta J(\theta)=-\nabla_\theta\mathcal L(\theta)$.

\paragraph{Sufficient conditions.}
We impose the following conditions.

\begin{enumerate}
    \item \textbf{Fixed decision procedure.}
    Responses have a bounded maximum length and a
    parameter-independent support.
    Parsing, including the handling of invalid responses,
    follows a fixed rule that maps each response to a
    valid parameter prediction.
    The downstream optimizer returns a measurable optimal
    decision in $\mathcal W$, with fixed tie-breaking.

    \item \textbf{Finite decision-regret moments.}
    There exists a measurable envelope $R(c)\ge 0$ such that, for every
    supported response $y$,
    \begin{equation}
        \ell(g(y),c) \le R(c),
        \label{eq:app_regret_envelope}
    \end{equation}
    and $\mathbb E_{(x,c)\sim\mathcal D}[R(c)^2]<\infty$.
    When $\mathcal W$ is compact and $F(\cdot,c)$ is continuous, one valid
    choice is
    $R(c)=\sup_{w\in\mathcal W}\left[F(w,c)-F(w^\star(c),c)\right]$.
    As in~\Cref{app:inventory_construction}, $\mathcal W$ and $F$ may depend on the context $x$; we
    suppress this dependence, in which case $R$ is a function of $(x,c)$.

    \item \textbf{Policy regularity.}
    On an open convex parameter region $\Theta$,
    $p_\theta(y\mid x)$ is positive on its support and
    twice continuously differentiable in $\theta$.
    For finite constants $S,H$, uniformly over
    $\theta\in\Theta$, supported responses, and almost
    every context,
    \begin{equation*}
        \|\nabla_\theta\log p_\theta(y\mid x)\|
        \leq S,
        \qquad
        \|\nabla_\theta^2\log p_\theta(y\mid x)\|_{\mathrm{op}}
        \leq H.
        \label{eq:app_policy_regularity}
    \end{equation*}
\end{enumerate}

The moment condition permits regret to vary across
instances and does not require a uniform bound on
the observed parameters.
For example, if $F(w,c)=c^\top w$ and $\mathcal W$
has finite diameter $D_{\mathcal W}$, then the supremum choice satisfies $R(c)\le D_{\mathcal W}\|c\|$.
Consequently, $\mathbb E[\|c\|^2]<\infty$ is sufficient.
The policy conditions concern the LLM rather than the
downstream optimization problem; they are explicit
regularity assumptions, not automatic consequences
of the softmax parameterization.

\paragraph{Verification of smoothness and finite variance.}
Write $s_\theta(y,x)=\nabla_\theta\log p_\theta(y\mid x)$.
Since parsing and reward evaluation do not depend
explicitly on $\theta$,
\begin{equation}
    \nabla_\theta^2 J(\theta)
    =
    \mathbb E_{(x,c)\sim\mathcal D,\,
               y\sim p_\theta(\cdot\mid x)}
    \left[
        r(y,c)
        \left(
            \nabla_\theta^2\log p_\theta(y\mid x)
            +
            s_\theta(y,x)s_\theta(y,x)^\top
        \right)
    \right].
    \label{eq:app_reward_hessian}
\end{equation}
The response support is finite, and the regret envelope
and derivative bounds justify differentiation under
the data expectation.
Because $|r(y,c)|\leq R(c)$,
Equation~\eqref{eq:app_reward_hessian} implies
\begin{equation}
    \|\nabla_\theta^2J(\theta)\|_{\mathrm{op}}
    \leq
    (H+S^2)\mathbb E[R(c)]
    =:L_J.
    \label{eq:app_objective_smoothness}
\end{equation}
Thus, both $J$ and $\mathcal L=-J$ have Lipschitz
gradients on $\Theta$, even when the optimizer's
solution map is discontinuous.

For a fixed instance $(x,c)$, all rewards lie in
$[-R(c),0]$.
The RLOO advantages therefore satisfy $|A_k|\leq R(c)$,
and the response-level estimator satisfies
$\|\widehat G_\theta^{\mathrm{RLOO}}(x,c)\|\leq SR(c)$.
Together with unbiasedness in
Equation~\eqref{eq:dfrl_rloo_unbiased}, this gives
\begin{equation}
    \mathbb E\left[
        \left\|
            \widehat G_\theta^{\mathrm{RLOO}}(x,c)
            -
            \nabla_\theta J(\theta)
        \right\|^2
    \right]
    \leq
    S^2\mathbb E[R(c)^2]
    =:\sigma^2.
    \label{eq:app_rloo_variance}
\end{equation}

\paragraph{Convergence consequence.}
Consider the stochastic-gradient update
\begin{equation}
    \theta_{t+1}
    =
    \theta_t+\eta\widehat G_t^{\mathrm{RLOO}},
    \label{eq:app_rloo_update}
\end{equation}
using a fresh pair $(x_t,c_t)\sim\mathcal D$ and
$K\geq 2$ conditionally independent responses from
the current policy at each iteration.
Assume that the iterates remain in $\Theta$.
For $0<\eta\leq 1/L_J$, standard stochastic-gradient
analysis~\citep{ghadimi2013stochastic}, together with
$\mathcal L(\theta)\geq 0$, yields
\begin{equation}
    \frac{1}{T}\sum_{t=0}^{T-1}
    \mathbb E\left[
        \|\nabla_\theta\mathcal L(\theta_t)\|^2
    \right]
    \leq
    \frac{2\mathcal L(\theta_0)}{\eta T}
    +
    L_J\eta\sigma^2.
    \label{eq:app_rloo_convergence}
\end{equation}
Choosing a step size proportional to $T^{-1/2}$,
subject to $\eta\leq 1/L_J$, gives an
$O(T^{-1/2})$ bound on the average expected squared
gradient norm.

Equation~\eqref{eq:app_rloo_convergence} establishes
approach to stationarity of expected decision regret
for the update in Equation~\eqref{eq:app_rloo_update}.
If $\mathcal D$ is the empirical training distribution,
the guarantee concerns the empirical expected-regret
objective.

\subsection{Analysis of Heuristic Warm-Starting}
\label[appendix]{app:heuristic_warm_starting}

We study a warm-start phase using inexpensive
rewards, followed by refinement using an unbiased
policy-gradient estimator.
Throughout, $\mathcal L(\theta)=-J(\theta)$ is the
exact expected-regret objective.

\paragraph{Assumptions and notation.}
We retain the assumptions in
\Cref{app:rloo_convergence}.
The heuristic is a fixed measurable mapping
$\widetilde w(\hat c)\in W$, with any tie-breaking
independent of $\theta$.
We assume that its realized regret
$F(\widetilde w(\hat c),c)-F(w^\star(c),c)$ is bounded by the same envelope
$R(c)$; this holds automatically for the supremum choice, since
$\widetilde w(\hat c)\in\mathcal W$.
Together with the score bound $S$, this ensures
finite second moments for its RLOO estimator.
Feasibility alone, however, does not ensure that
its gradient bias is small.

Let $\mathcal F_t$ denote the training history
before iteration $t$.
Each update uses a fresh batch and conditionally
independent policy samples.
For the warm-start phase, define
\[
    b_\theta
    :=
    \mathbb E[\widetilde G_\theta]
    -\nabla_\theta J(\theta),
    \qquad
    v_{\mathrm h}(\theta)
    :=
    \mathbb E\left[
        \|\widetilde G_\theta
          -\mathbb E[\widetilde G_\theta]\|^2
    \right].
\]
Assume that, throughout this phase,
\[
    \|b_\theta\|\leq\varepsilon_{\mathrm h},
    \qquad
    v_{\mathrm h}(\theta)\leq\sigma_{\mathrm h}^2.
\]
All iterates and the line segments between
successive iterates are assumed to remain in
the parameter region $\Theta$.

\paragraph{When is the bias small?}
Using the exact and inexpensive batch estimators
in Equation~\eqref{eq:dfrl_estimated_gradient},
unbiasedness of $G_\theta$ gives
\[
    b_\theta
    =
    \mathbb E[\widetilde G_\theta-G_\theta].
\]
Jensen's inequality therefore yields
\begin{equation}
    \|b_\theta\|^2
    \leq
    \mathbb E[
        \|\widetilde G_\theta-G_\theta\|^2
    ]
    =
    D_\theta,
    \label{eq:heuristic_bias_gradient_error}
\end{equation}
where $D_\theta$ is the approximation-error
quantity in \Cref{app:correction_efficiency}.
Thus, accurate approximation of the exact batch
gradient is sufficient for small warm-start bias.

A complementary condition can be stated in
terms of realized decision costs.
For one prediction, write
\[
    \delta(x,c,\hat c)
    :=
    F(\widetilde w(\hat c),c)
    -
    F(w^\star(\hat c),c).
\]
Because the leave-one-out baseline is independent
of the current prediction conditional on $(x,c)$,
its contribution to the expected score-weighted
cost difference is zero.
In the response-level notation of
\Cref{app:rloo_convergence}, this gives
\[
    b_\theta
    =
    -\mathbb E\left[
        \delta(x,c,g(y))s_\theta(y,x)
    \right].
\]
The score bound and Cauchy--Schwarz imply
\begin{equation}
    \|b_\theta\|^2
    \leq
    S^2\,
    \mathbb E[
        \delta(x,c,g(y))^2
    ].
    \label{eq:heuristic_bias_cost_error}
\end{equation}
Hence, small mean-squared realized-cost error
also provides a sufficient condition.
An approximation guarantee under the predicted
parameters $\hat c$ alone does not necessarily
control this error, which is evaluated under
the observed parameters $c$.

\paragraph{Descent during warm-starting.}
Consider
\[
    \theta_{t+1}
    =
    \theta_t
    +\eta_{\mathrm h}\widetilde G_{\theta_t},
    \qquad
    0<\eta_{\mathrm h}\leq 1/L_J.
\]
Write
$g_t:=\nabla_\theta\mathcal L(\theta_t)$
and $b_t:=b_{\theta_t}$.
Since $\nabla_\theta J=-\nabla_\theta\mathcal L$,
\[
    \mathbb E[
        \widetilde G_{\theta_t}\mid\mathcal F_t
    ]
    =
    -g_t+b_t.
\]
Smoothness gives
\begin{align*}
    \mathbb E[
        \mathcal L(\theta_{t+1})
        \mid\mathcal F_t
    ]
    &\leq
    \mathcal L(\theta_t)
    +\eta_{\mathrm h}
        \langle g_t,-g_t+b_t\rangle
    \nonumber\\
    &\quad+
    \frac{L_J\eta_{\mathrm h}^2}{2}
    \left(
        \|-g_t+b_t\|^2+\sigma_{\mathrm h}^2
    \right).
    \label{eq:heuristic_smoothness_step}
\end{align*}
Using
\[
    2\langle g_t,-g_t+b_t\rangle
    =
    \|b_t\|^2-\|g_t\|^2-\|-g_t+b_t\|^2
\]
and $L_J\eta_{\mathrm h}\leq 1$, we obtain
\[
    \mathbb E[
        \mathcal L(\theta_{t+1})
        \mid\mathcal F_t
    ]
    \leq
    \mathcal L(\theta_t)
    -
    \frac{\eta_{\mathrm h}}{2}\|g_t\|^2
    +
    \frac{\eta_{\mathrm h}}{2}
        \varepsilon_{\mathrm h}^2
    +
    \frac{L_J\eta_{\mathrm h}^2}{2}
        \sigma_{\mathrm h}^2.
\]
This proves
Equation~\eqref{eq:heuristic_warm_descent}.

In particular, a sufficient condition for strict
expected descent is
\[
    \|g_t\|^2
    >
    \varepsilon_{\mathrm h}^2
    +L_J\eta_{\mathrm h}\sigma_{\mathrm h}^2.
\]
Alternatively, if
$\|b_t\|\leq\rho\|g_t\|$ for some $\rho<1$,
strict expected descent follows whenever
\[
    (1-\rho^2)\|g_t\|^2
    >
    L_J\eta_{\mathrm h}\sigma_{\mathrm h}^2.
\]
These are sufficient conditions; they need not
hold throughout training.

Summing over $T_{\mathrm w}$ warm-start updates
and using $\mathcal L\geq 0$ yields
\begin{equation}
    \frac{1}{T_{\mathrm w}}
    \sum_{t=0}^{T_{\mathrm w}-1}
    \mathbb E[
        \|\nabla_\theta\mathcal L(\theta_t)\|^2
    ]
    \leq
    \frac{2\mathcal L(\theta_0)}
         {\eta_{\mathrm h}T_{\mathrm w}}
    +
    \varepsilon_{\mathrm h}^2
    +
    L_J\eta_{\mathrm h}\sigma_{\mathrm h}^2.
    \label{eq:heuristic_warm_stationarity}
\end{equation}
The bias term does not vanish by reducing the
step size or increasing the number of updates.
This limits the guarantee available from
heuristic-only training and motivates switching
to an unbiased estimator.

\paragraph{Refinement after warm-starting.}
After $T_{\mathrm w}$ updates, use an estimator
$\widehat G_t^{\mathrm{ref}}$ satisfying
\[
    \mathbb E[
        \widehat G_t^{\mathrm{ref}}\mid\mathcal F_t
    ]
    =
    \nabla_\theta J(\theta_t),
    \qquad
    \mathbb E\left[
        \|\widehat G_t^{\mathrm{ref}}
          -\nabla_\theta J(\theta_t)\|^2
        \mid\mathcal F_t
    \right]
    \leq\sigma_{\mathrm{ref}}^2.
\]
Exact RLOO satisfies this condition under
the assumptions in \Cref{app:rloo_convergence}.

With $T_{\mathrm r}$ refinement updates and
$0<\eta_{\mathrm r}\leq 1/L_J$, the same
smoothness argument gives
\begin{equation}
    \frac{1}{T_{\mathrm r}}
    \sum_{t=T_{\mathrm w}}^{T_{\mathrm w}
                            +T_{\mathrm r}-1}
    \mathbb E[
        \|\nabla_\theta\mathcal L(\theta_t)\|^2
    ]
    \leq
    \frac{
        2\mathbb E[\mathcal L(\theta_{T_{\mathrm w}})]
    }{\eta_{\mathrm r}T_{\mathrm r}}
    +
    L_J\eta_{\mathrm r}\sigma_{\mathrm{ref}}^2.
    \label{eq:heuristic_refinement_bound}
\end{equation}
Thus, warm-starting affects this bound through
the objective value at the switch.
Its gradient bias does not appear as a persistent
bias term during unbiased refinement.

\paragraph{Accounting for the time budget.}
Let
\[
    \tau_{\mathrm{warm}}
    :=
    \tau_0+\tau_{\mathrm{cheap}},
    \qquad
    \tau_{\mathrm{full}}
    :=
    \tau_0+\tau_{\mathrm{exact}},
\]
using the timing notation in
\Cref{app:correction_efficiency}.
For this comparison, take these quantities as
uniform upper bounds on the conditional expected
batch times in the respective phases.
Fixed update counts satisfying
\[
    T_{\mathrm w}\tau_{\mathrm{warm}}
    +
    T_{\mathrm r}\tau_{\mathrm{full}}
    \leq B
\]
give a warm-start--then-exact procedure within
expected budget $B$.

An exact-only procedure can instead perform
$T_{\mathrm{full}}
:=\lfloor B/\tau_{\mathrm{full}}\rfloor$
updates.
For positive $T_{\mathrm r}$ and
$T_{\mathrm{full}}$, the same refinement step
size, and a common exact-gradient variance bound,
the warm-start procedure has a smaller
stationarity upper bound whenever
\begin{equation}
    \mathbb E[
        \mathcal L(\theta_{T_{\mathrm w}})
    ]
    <
    \frac{T_{\mathrm r}}{T_{\mathrm{full}}}
    \mathcal L(\theta_0).
    \label{eq:heuristic_budget_condition}
\end{equation}
This condition requires the reduction in the
objective during warm-starting to compensate
for the smaller number of exact refinement
updates.
It compares theoretical upper bounds and does
not establish an ordering of the realized
training trajectories or final test regret.

The analysis concerns stochastic-gradient
updates and expected sampled-policy regret.
The behavior of Adam and the regret obtained
from deterministic mean predictions are
evaluated empirically.

\subsection{Correction Efficiency under a Computational Budget}
\label[appendix]{app:correction_efficiency}

We compare the exact and corrected batch estimators
under a fixed expected time budget.
The analysis accounts for the additional variance
introduced by correction and the time saved when
an entire batch skips exact reward evaluation.

\paragraph{Variance of the corrected batch estimator.}
At a fixed $\theta$, define
\begin{equation}
    V_\theta
    :=
    \mathbb E\!\left[
        \|G_\theta-\nabla_\theta J(\theta)\|^2
    \right],
    \qquad
    D_\theta
    :=
    \mathbb E\!\left[
        \|G_\theta-\widetilde G_\theta\|^2
    \right].
    \label{eq:dfrl_variance_error}
\end{equation}
Both quantities refer to batch averages:
$V_\theta$ is the variance of the exact batch estimator,
and $D_\theta$ is the mean-squared error of its
inexpensive approximation.

Equation~\eqref{eq:dfrl_corrected_gradient} can be written as
\begin{equation*}
    \widehat G_\theta^{\mathrm{corr}}
    =
    G_\theta
    +
    \left(\frac{I}{p}-1\right)
    (G_\theta-\widetilde G_\theta).
    \label{eq:dfrl_correction_decomposition}
\end{equation*}
Conditional on the batch, predictions, and cost estimates,
the correction error has zero mean and second moment
\[
    \mathbb E_I\!\left[
        \|\widehat G_\theta^{\mathrm{corr}}-G_\theta\|^2
    \right]
    =
    \frac{1-p}{p}
    \|G_\theta-\widetilde G_\theta\|^2.
\]
The cross term with
$G_\theta-\nabla_\theta J(\theta)$ therefore vanishes
in expectation, giving
\begin{equation}
    V_\theta^{\mathrm{corr}}(p)
    :=
    \mathbb E\!\left[
        \|\widehat G_\theta^{\mathrm{corr}}
        -\nabla_\theta J(\theta)\|^2
    \right]
    =
    V_\theta+\frac{1-p}{p}D_\theta.
    \label{eq:dfrl_corrected_variance}
\end{equation}
For $p<1$, Equation~\eqref{eq:dfrl_corrected_variance}
shows that the correction adds variance
$(1-p)D_\theta/p$ to the exact batch estimator.
It can nevertheless improve gradient estimation under
a fixed time budget if faster batch processing allows
enough additional estimates to offset this increase.

\paragraph{Expected batch processing time.}
Let $T_{\mathrm{exact}}$ and $T_{\mathrm{cheap}}$ denote
the times required to compute all $MK$ exact decision
costs and their inexpensive estimates, respectively.
Assuming each phase evaluates the predictions independently
with sufficient parallel resources,
\[
    T_{\mathrm{exact}}
    =
    \max_{\substack{1\leq m\leq M\\1\leq k\leq K}}
    t_{mk}^{\mathrm{exact}},
    \qquad
    T_{\mathrm{cheap}}
    =
    \max_{\substack{1\leq m\leq M\\1\leq k\leq K}}
    t_{mk}^{\mathrm{cheap}},
\]
where $t_{mk}^{\mathrm{exact}}$ and $t_{mk}^{\mathrm{cheap}}$
are the times required to compute $C_{mk}$ and
$\widetilde C_{mk}$, respectively.
Define
\[
    \tau_{\mathrm{exact}}
    :=\mathbb E[T_{\mathrm{exact}}],
    \qquad
    \tau_{\mathrm{cheap}}
    :=\mathbb E[T_{\mathrm{cheap}}].
\]
These expectations account for runtime variation across
sampled batches.
Let $T_0$ denote the time for prediction generation
and the model update, with $\tau_0:=\mathbb E[T_0]$.
The total batch processing time is
$T_{\mathrm{batch}}=T_0+T_{\mathrm{cheap}}+I T_{\mathrm{exact}}$.
Therefore,
\begin{equation}
    \tau_{\mathrm{batch}}(p)
    :=
    \mathbb E[T_{\mathrm{batch}}]
    =
    \tau_0+\tau_{\mathrm{cheap}}
    +p\tau_{\mathrm{exact}}.
    \label{eq:dfrl_parallel_cost}
\end{equation}

Exact RLOO uses the same execution schedule
without inexpensive reward evaluation, giving
\begin{equation*}
    \tau_{\mathrm{full}}
    :=
    \tau_0+\tau_{\mathrm{exact}}.
    \label{eq:dfrl_full_batch_cost}
\end{equation*}
Because selection operates on the whole batch,
the exact-evaluation phase is skipped with probability
$1-p$.
The expected number of exact solves per batch is $pMK$.

\paragraph{Gradient accuracy under a time budget.}
For an expected time budget $\mathcal B$, define
\[
    N_{\mathrm{corr}}
    :=
    \left\lfloor
        \frac{\mathcal B}{\tau_{\mathrm{batch}}(p)}
    \right\rfloor,
    \qquad
    N_{\mathrm{full}}
    :=
    \left\lfloor
        \frac{\mathcal B}{\tau_{\mathrm{full}}}
    \right\rfloor.
\]
At a fixed policy, averaging independently generated
batch estimators gives, for positive batch counts,
\begin{equation*}
    \operatorname{MSE}_{\mathrm{corr}}
    =
    \frac{V_\theta+(1-p)D_\theta/p}
         {N_{\mathrm{corr}}},
    \qquad
    \operatorname{MSE}_{\mathrm{full}}
    =
    \frac{V_\theta}{N_{\mathrm{full}}}.
    \label{eq:dfrl_parallel_mse}
\end{equation*}
Ignoring integer rounding, the correction improves
gradient accuracy under the same expected time budget
precisely when
\begin{equation}
    \tau_{\mathrm{batch}}(p)
    \left(
        V_\theta+\frac{1-p}{p}D_\theta
    \right)
    <
    \tau_{\mathrm{full}}V_\theta.
    \label{eq:dfrl_parallel_efficiency}
\end{equation}
For $V_\theta>0$ and $0<p<1$, this is equivalent to
\begin{equation}
    \frac{D_\theta}{V_\theta}
    <
    \frac{p}{1-p}
    \left(
        \frac{\tau_{\mathrm{full}}}
             {\tau_{\mathrm{batch}}(p)}
        -1
    \right).
    \label{eq:dfrl_error_threshold}
\end{equation}
Thus, the time saved must compensate for the additional
variance.
A necessary condition is
$\tau_{\mathrm{batch}}(p)<\tau_{\mathrm{full}}$,
which becomes
\begin{equation*}
    \tau_{\mathrm{cheap}}
    <
    (1-p)\tau_{\mathrm{exact}}.
    \label{eq:dfrl_time_saving_condition}
\end{equation*}
The inexpensive evaluation time must therefore be smaller
than the expected exact-evaluation time saved.

\paragraph{Selecting the evaluation probability.}
Using $\mathcal B/\tau_{\mathrm{batch}}(p)$ to approximate
the number of processed batches, the mean-squared error
is $\Phi_\theta(p)/\mathcal B$, where
\begin{equation*}
    \Phi_\theta(p)
    :=
    \left(
        \tau_0+\tau_{\mathrm{cheap}}
        +p\tau_{\mathrm{exact}}
    \right)
    \left(
        V_\theta+\frac{1-p}{p}D_\theta
    \right),
    \qquad 0<p\leq 1.
    \label{eq:dfrl_probability_selection}
\end{equation*}
For $0<D_\theta<V_\theta$ and positive timing constants,
the minimizer is
\begin{equation}
    p^\star
    =
    \min\left\{
        1,\,
        \sqrt{
            \frac{
                (\tau_0+\tau_{\mathrm{cheap}})D_\theta
            }{
                \tau_{\mathrm{exact}}
                (V_\theta-D_\theta)
            }
        }
    \right\}.
    \label{eq:dfrl_optimal_probability}
\end{equation}
This choice balances the additional variance against
the time saved by skipping exact evaluation.
It improves over exact RLOO only if
$\Phi_\theta(p^\star)<\tau_{\mathrm{full}}V_\theta$.

If $D_\theta\geq V_\theta>0$, the minimum occurs at
$p=1$, which reproduces the exact estimator while
incurring the additional estimation time.
If $D_\theta=0$, reducing $p$ does not increase variance.
These are fixed-policy comparisons; selecting $p$
in practice requires estimates of the timing and
gradient-error quantities, which may change during
training.

\paragraph{Implications for convergence.}
Suppose that, conditional on the training history,
the exact batch variance and approximation error satisfy
$V_{\theta_t}\leq\overline V$ and
$D_{\theta_t}\leq\overline D$ throughout training.
For a fixed $p>0$, Equation~\eqref{eq:dfrl_corrected_variance}
gives the batch variance bound
\[
    \sigma_p^2
    :=
    \overline V+\frac{1-p}{p}\overline D.
\]
Under the remaining assumptions in
\Cref{app:rloo_convergence}, fresh-batch updates using
$\widehat G_{\theta_t}^{\mathrm{corr}}$ satisfy
\begin{equation}
    \frac1T\sum_{t=0}^{T-1}
    \mathbb E\!\left[
        \|\nabla_\theta\mathcal L(\theta_t)\|^2
    \right]
    \leq
    \frac{2\mathcal L(\theta_0)}{\eta T}
    +
    L_J\eta
    \left(
        \overline V+\frac{1-p}{p}\overline D
    \right),
    \qquad 0<\eta\leq 1/L_J.
    \label{eq:dfrl_corrected_convergence}
\end{equation}
If Equation~\eqref{eq:dfrl_parallel_cost} bounds the
expected batch time throughout training, then
$T=\lfloor\mathcal B/\tau_{\mathrm{batch}}(p)\rfloor$
updates satisfy the time budget in expectation.
With an appropriate step size, the resulting
stationarity bound has order
\begin{equation}
    O\left(
        \sqrt{
            \frac{\tau_{\mathrm{batch}}(p)}{\mathcal B}
            \left(
                \overline V
                +\frac{1-p}{p}\overline D
            \right)
        }
        +
        \frac{\tau_{\mathrm{batch}}(p)}{\mathcal B}
    \right).
    \label{eq:dfrl_parallel_convergence}
\end{equation}
The leading term improves over exact RLOO when
Equation~\eqref{eq:dfrl_parallel_efficiency} holds with
$\overline V$ and $\overline D$.
The correction therefore preserves the stationarity
guarantee, while its computational benefit depends
on whether the batch processing time saved outweighs
the additional estimator variance.

\subsection{SFT as Prediction-Loss Minimization}
\label[appendix]{app:sft_prediction_loss}

SFT can be understood as probabilistic prediction using
log loss, with numerical targets represented as text.
Let $q(c\mid x)$ denote the true conditional distribution
of the observed parameters.
A probabilistic predictor $\pi_\theta$ trained by log loss
minimizes
\begin{equation}
    \mathcal L_{\mathrm{pred}}(\theta)
    :=
    \mathbb E_{(x,c)\sim\mathcal D}
    [-\log\pi_\theta(c\mid x)].
    \label{eq:app_prediction_log_loss}
\end{equation}
For discrete targets, this objective decomposes as
\begin{equation*}
    \mathcal L_{\mathrm{pred}}(\theta)
    =
    \mathbb E_x
    \left[
        H(q(\cdot\mid x))
        +
        D_{\mathrm{KL}}
        \bigl(q(\cdot\mid x)\,\|\,\pi_\theta(\cdot\mid x)\bigr)
    \right].
    \label{eq:app_prediction_kl}
\end{equation*}
Consequently, when the model class can represent $q$,
the population optimum is
$\pi_\theta(\cdot\mid x)=q(\cdot\mid x)$
for almost every $x$.

For an LLM, let $y(c)$ denote a fixed, uniquely decodable
token sequence representing $c$, and let
$p_\theta(\cdot\mid x)$ denote the response distribution.
The sequence-level negative log-likelihood is
\begin{equation}
    -\log p_\theta(y(c)\mid x)
    =
    -\sum_{t=1}^{T(c)}
    \log p_\theta(y_t(c)\mid x,y_{<t}(c)).
    \label{eq:app_sft_sequence_loss}
\end{equation}
Let $q_Y(\cdot\mid x)$ be the distribution obtained by
encoding $c\sim q(\cdot\mid x)$ as $y(c)$.
The expected loss in
Equation~\eqref{eq:app_sft_sequence_loss} similarly equals
\begin{equation*}
    \mathbb E_x
    \left[
        H(q_Y(\cdot\mid x))
        +
        D_{\mathrm{KL}}
        \bigl(q_Y(\cdot\mid x)\,\|\,p_\theta(\cdot\mid x)\bigr)
    \right].
    \label{eq:app_sft_kl}
\end{equation*}
Its population optimum is therefore $p_\theta=q_Y$,
provided this distribution is representable.
Parsing responses from this optimal model recovers
$q(c\mid x)$, the same optimal parameter distribution
as prediction with log loss in
Equation~\eqref{eq:app_prediction_log_loss}.

\section{Experimental Details and Additional Results}
\label[appendix]{app:experimental details and additional results}

\subsection{Synthetic Problems and Data Generation}
\label[appendix]{app:synthetic_construction}

\paragraph{Shortest path.}
We use a $5\times5$ directed grid with rightward and
downward edges, giving 40 edges and 70 source-to-target
paths.
The unknown parameters $c\in\mathbb R^{40}$ are edge
costs. The optimization problem is
\[
    \min_{w\in[0,1]^{40}} c^\top w
    \quad\text{subject to}\quad
    A_{\mathrm{flow}}w=b_{\mathrm{flow}},
\]
where $b_{\mathrm{flow}}$ specifies one unit of flow
from the source to the target.

\paragraph{Knapsack.}
We consider 12 binary item-selection variables and two
resource constraints:
\[
    \min_{w\in\{0,1\}^{12}} c^\top w
    \quad\text{subject to}\quad
    Mw\leq b.
\]
Each entry of $M\in\mathbb R^{2\times12}$ is sampled
uniformly from $\{3.00,3.01,\ldots,7.99\}$.
Each resource capacity equals 35\% of the sum of
the corresponding row of $M$.
The coefficients $c$ represent negative item profits.

\paragraph{Quadratic allocation.}
We consider the strictly convex problem
\[
    \min_{w\in\mathbb R_+^{12}}
    c^\top w+\frac12\|w\|_2^2
    \quad\text{subject to}\quad
    \mathbf 1^\top w=3.
\]
For the finite-decision problems, ties are resolved
using a fixed ordering independent of the observed
parameters.

\paragraph{Parameter generation.}
For each problem, let $d$ denote the parameter dimension
and draw a fixed matrix $B\in\{0,1\}^{d\times5}$ with
independent $\mathrm{Bernoulli}(0.5)$ entries.
For each instance, independently sample
$x\sim\mathcal N(0,I_5)$ and
$\varepsilon_j\sim\mathrm{Unif}[0.5,1.5]$.
Define
\begin{equation*}
    s_j(x)
    :=
    \frac{
        \left((Bx)_j/\sqrt{5}+3\right)^4+1
    }{3.5^4}.
    \label{eq:app_synthetic_signal}
\end{equation*}
The optimization parameters are
\begin{equation*}
    c_j
    =
    \begin{cases}
        s_j(x)\varepsilon_j,
        & \text{shortest path and allocation},\\
        -\lceil5s_j(x)\varepsilon_j\rceil,
        & \text{knapsack}.
    \end{cases}
    \label{eq:app_synthetic_parameters}
\end{equation*}
We use data-generation seed 135 and retain one dataset
per problem, split into 1,000 training, 250 validation,
and 10,000 test instances.

\subsection{Synthetic Training and Evaluation Details}
\label[appendix]{app:synthetic_training}

\paragraph{Predictors and prediction-loss baselines.}
All methods use an affine predictor
$\mu_\theta(x)=W_\theta x+b_\theta$,
with inputs in their original coordinates.
For DFRL, $\mu_\theta(x)$ specifies
the Gaussian mean.
The squared-error and absolute-error baselines minimize
their empirical prediction losses with an $\ell_1$
penalty on $W_\theta$; intercepts are unpenalized.
The squared-error loss includes a factor of $1/2$.
Regularization strengths are selected by validation
decision regret from ten logarithmically spaced values
between $10^{-6}$ and $10^2$.

\paragraph{SPO+ baseline.}
We include SPO+ for shortest path and knapsack,
whose objectives are linear in the predicted parameters.
The predictor minimizes the regularized empirical
SPO+ surrogate, using the same regularization grid
and selection criterion as the prediction-loss baselines.
We do not include a SPO+ baseline for quadratic
allocation and mark the corresponding entries as N/A.
Table~\ref{tab:synthetic_regularization} reports the
selected regularization strengths.

\begin{table}[htbp]
\centering
\caption{Selected regularization strengths for the
synthetic baselines. SPO+ is evaluated only on
shortest path and knapsack.}
\label{tab:synthetic_regularization}
\small
\setlength{\tabcolsep}{5pt}
\renewcommand{\arraystretch}{1.18}
\begin{tabular}{lrrr}
\toprule
Method & Shortest path & Knapsack & Allocation \\
\midrule
Squared error
& $10^{-6}$
& $2.7826\times10^{-2}$
& $2.7826\times10^{-2}$ \\
Absolute error
& $2.7826\times10^{-2}$
& $4.6416\times10^{-4}$
& $2.7826\times10^{-2}$ \\
SPO+
& $4.6416\times10^{-4}$
& $2.7826\times10^{-2}$
& N/A \\
\bottomrule
\end{tabular}
\end{table}

\paragraph{DFRL with RLOO.}
Let
\[
    s_c
    :=
    \frac{1}{N_{\mathrm{train}}d}
    \sum_{i=1}^{N_{\mathrm{train}}}\|c_i\|_1
\]
denote the training mean absolute parameter value.
DFRL uses the Gaussian policy
\[
    \pi_\theta(\cdot\mid x)
    =
    \mathcal N\!\left(\mu_\theta(x),(0.1s_c)^2I\right),
\]
initialized from the selected squared-error predictor.
Each update samples 16 distinct training contexts
and eight independent predictions per context.
Advantages subtract the leave-one-out baseline without
minibatch standardization.
The normalized learning rate is 0.001.

\paragraph{Optimization and checkpoint selection.}
DFRL uses Adam with
$(\beta_1,\beta_2)=(0.9,0.999)$ and
$\epsilon=10^{-8}$, without gradient clipping.
The mean-parameter learning rate equals the normalized
learning rate multiplied by $s_c$.
Validation is performed every 32,000 reward evaluations.
For each run, we select the checkpoint whose mean
predictions induce the lowest validation regret,
including initialization as a candidate.
Final training uses seeds 235, 236, and 237.

\paragraph{Evaluation.}
For DFRL, we report the mean
and sample standard deviation across the three
training seeds.
These standard deviations measure training variability
conditional on the fixed dataset.
We additionally estimate expected decision regret
using 16 independent Gaussian predictions per test
instance, as reported in
Table~\ref{tab:synthetic_sampled_results}.
For the deterministic prediction-loss and SPO+
baselines, we add Gaussian noise with reference
covariance $(0.1s_c)^2I$.
These baseline rows therefore evaluate stochastic
extensions of the trained point predictors.
We use common standard-normal draws across methods,
with evaluation seed 981531.

\begin{table}[htbp]
\centering
\caption{Sampled test regret, averaged over 16 predictions
per test instance. DFRL uses its own
covariance; deterministic baselines use the reference
covariance $(0.1s_c)^2I$.
Values following $\pm$ are sample standard deviations
over three training seeds.
Lower is better; bold values indicate the lowest mean
in each column.}
\label{tab:synthetic_sampled_results}
\small
\setlength{\tabcolsep}{5pt}
\renewcommand{\arraystretch}{1.18}
\begin{tabular}{lrrr}
\toprule
Method & Shortest path & Knapsack & Allocation \\
\midrule
Squared-error prediction
& 0.3402 & 1.5341 & 0.1993 \\
Absolute-error prediction
& 0.3326 & 1.5828 & 0.1806 \\
SPO+
& 0.2972 & 1.3845 & N/A \\
\rowcolor{msftcard}
\textbf{DFRL (RLOO)}
& $\mathbf{0.2882}\pm0.0025$
& $\mathbf{1.3337}\pm0.0040$
& $\mathbf{0.1721}\pm0.0004$ \\
\bottomrule
\end{tabular}
\end{table}

\subsection{Additional Details for the Computational
Efficiency Experiments}
\label[appendix]{app:efficiency_experiments}

\subsubsection{Problem Setup and Training Configuration}

We consider a 96-item knapsack problem with two
resource constraints:
\[
    \min_{w\in\{0,1\}^{96}} c^\top w
    \qquad
    \text{subject to }
    \mathbf{M}w\leq b.
\]
Costs are represented as negative profits.
The resource matrix $\mathbf{M}\in\mathbb{R}^{2\times96}$
is fixed across instances, with entries sampled from
$\{3.00,3.01,\ldots,7.99\}$.
Each capacity is set to $35\%$ of the corresponding
row sum.
Here, $\mathbf{M}$ denotes the resource matrix,
whereas $M$ below denotes the training batch size.

We use the PyEPO knapsack data generator with
five context features, polynomial degree four,
multiplicative noise width $0.5$, and data seed 135.
The training, validation, and test sets contain
1,000, 250, and 10,000 instances, respectively.
All DFRL methods use the same affine Gaussian policy
\[
    \pi_\theta(\cdot\mid x)
    =
    \mathcal{N}\!\left(
        \mu_\theta(x),\sigma^2 I_{96}
    \right),
\]
where $\mu_\theta(x)$ includes an intercept and is
initialized from the same least-squares predictor.
The covariance remains fixed during training.

Each training batch contains $M=16$
distinct training instances and $K=8$ independent
Gaussian predictions per instance.
Let $s_c$ denote the mean absolute cost coefficient
over the training set.
For this dataset, $s_c=4.3516667$.
We use $\sigma=0.1s_c$ and divide realized costs
by $96s_c$ before constructing the within-instance
RLOO advantages.
This common normalization is used across methods;
reported test regrets remain in the original units.

Adam uses $(\beta_1,\beta_2)=(0.9,0.999)$,
numerical constant $10^{-8}$, zero initial moments,
and no gradient clipping or weight decay.
The nominal learning rate follows
\[
    \eta_{\mathrm{nom}}(t)
    =
    10^{-4}
    +
    \frac{10^{-3}-10^{-4}}{2}
    \left[
        1+\cos\!\left(\frac{\pi t}{T}\right)
    \right],
    \qquad 0\leq t\leq T,
\]
where $t$ is elapsed training time and $T=3600$
seconds.
Under the implementation's cost-scaling convention,
the learning rate passed to Adam is
$s_c\eta_{\mathrm{nom}}(t)$.
Its endpoints are therefore approximately
$0.004352$ and $0.0004352$.
All methods share this prescribed schedule;
learning rates are not tuned separately for
each evaluation probability.
Final training seeds are 435, 436, and 437,
paired across methods.

\subsubsection{Heuristic Feedback and Warm-Starting}

\paragraph{Inexpensive decision heuristic.}
For each item $j$, define its normalized resource
consumption as
\[
    d_j
    :=
    \sum_{r=1}^{2}\frac{\mathbf{M}_{rj}}{b_r}.
\]
Given a sampled prediction $\hat c$, the heuristic
sorts items by increasing $\hat c_j/d_j$,
breaking ties by item index.
Starting from an empty selection, it visits
items in this order and accepts an item when
$\hat c_j<0$ and the item fits within both
remaining capacities.
The feasibility check uses tolerance $10^{-10}$.
The resulting decision is denoted
$\widetilde w(\hat c)$.

The inexpensive realized cost is
\[
    \widetilde C
    =
    F(\widetilde w(\hat c),c)
    =
    c^\top\widetilde w(\hat c).
\]
Thus, predicted costs determine the heuristic
decision, while observed costs determine its
training feedback.
The heuristic produces a feasible decision
but is not assumed to recover the exact optimizer.

\paragraph{Reward correction.}
We apply the batch-level correction in
Equation~\eqref{eq:dfrl_corrected_advantage},
using the heuristic costs to construct
$\widetilde A_{mk}$.
A single independent Bernoulli draw selects
whether all $MK=128$ predictions in the batch
are evaluated exactly.
We test
$p\in\{0.05,0.1,0.25,0.5,0.75,0.9\}$.
No calibration coefficient is fitted.

\paragraph{Warm-start selection.}
We compare nominal switch times of 10, 60, and 300 seconds before
exact RLOO. Each candidate receives a full one-hour training run
with tuning seed 335. The switch time is selected by validation regret at
the one-hour checkpoint, with earlier switches preferred in a tie. This
procedure selects 10 seconds.

\subsubsection{Additional Results}

\paragraph{Update and exact-evaluation counts.}
Table~\ref{tab:efficiency_accounting}
reports the computational accounting for all
DFRL configurations.
An update denotes one Adam optimizer step.
Every training batch produces one update
in these experiments.
Exact-solver calls count individual optimization
problems, including solver-pool warm-up calls,
but exclude offline validation and test evaluation.
The warm-started method includes both stages
in its update counts.

Reducing $p$ decreases the probability of
exact evaluation per batch.
However, it also permits more batches within
the time budget, so the total number of
exact evaluations need not decrease.
For example, correction with $p=0.5$
performs approximately twice as many updates
as exact RLOO while making a similar number
of exact solver calls.
The time-budget comparison therefore measures
the combined effect of update frequency,
reward approximation, and correction.

\begin{table}[htbp]
\centering
\caption{Complete one-hour computational accounting
for the 96-item knapsack experiment.
Values are means $\pm$ sample standard deviations
across three paired training seeds.
Updates are in thousands, and training exact-solver
calls are in millions.
Counts correspond to the evaluated one-hour
checkpoints and include warm-start updates
and solver-pool warm-up calls.
Regret uses all 10,000 test instances.
The green row identifies heuristic warm-starting
followed by exact RLOO.
Bold identifies the lowest observed mean regret,
without implying statistical significance.}
\label{tab:efficiency_accounting}

\small
\setlength{\tabcolsep}{5pt}
\renewcommand{\arraystretch}{1.18}
\setlength{\aboverulesep}{0pt}
\setlength{\belowrulesep}{0pt}

\begin{tabular}{lrrr}
\toprule
Method
& Updates ($10^3$)
& Exact calls ($10^6$)
& Regret$\downarrow$ \\
\midrule
Exact RLOO
& $8.862\pm0.249$
& $1.134\pm0.032$
& $11.3669\pm0.0294$ \\
Heuristic-only
& $1380.172\pm16.778$
& $0$
& $11.9276\pm0.0280$ \\
\midrule
Correction, $p=0.05$
& $190.770\pm8.695$
& $1.219\pm0.057$
& $11.6653\pm0.0380$ \\
Correction, $p=0.1$
& $104.951\pm0.756$
& $1.330\pm0.019$
& $11.5619\pm0.0346$ \\
Correction, $p=0.25$
& $36.291\pm0.652$
& $1.156\pm0.022$
& $11.5038\pm0.0914$ \\
Correction, $p=0.5$
& $18.063\pm0.823$
& $1.158\pm0.037$
& $11.4437\pm0.0248$ \\
Correction, $p=0.75$
& $12.068\pm0.384$
& $1.162\pm0.029$
& $11.3737\pm0.0227$ \\
Correction, $p=0.9$
& $10.267\pm0.281$
& $1.186\pm0.036$
& $\mathbf{11.3558}\pm0.0103$ \\
\midrule
\rowcolor{msftcard}
\textbf{Heuristic $\to$ exact RLOO}
& $13.318\pm0.435$
& $1.212\pm0.041$
& $11.3690\pm0.0198$ \\
\bottomrule
\end{tabular}
\end{table}

\subsection{Inventory Dataset Construction}
\label[appendix]{app:inventory_construction}

\paragraph{Transactions and temporal splits.}
We aggregate transactions across sales channels into
daily product-level purchase counts, retaining duplicate
transaction rows as multiple purchased units.
Days without purchases are assigned zero sales.
We retain products with nonempty normalized descriptions.
Variants sharing a product code are grouped together;
families sharing identical normalized descriptions of at
least 40 characters are additionally merged for splitting.
The resulting groups are assigned to cohorts according
to their earliest observed purchase.

Training cohorts first appear through December 2019,
validation cohorts from January through March 2020,
and test cohorts from April through September 2020.
Groups first observed within the initial 90 days of the
transaction dataset are excluded.
Forecast origins are Mondays. Historical inputs use the
preceding seven days, and targets use the following
seven days. Target windows must fall entirely within
the corresponding split and the observed data period.

At each origin, eligible products have an observed age
of 7--35 days and positive sales during the preceding
week. Zero sales in the target week are retained.
We select one variant per product family using a seeded
hash and partition the selected products into groups
of 16, discarding incomplete groups.
All preprocessing parameters are fitted on training
data only.

\paragraph{Inventory parameters.}
For product $j$, let $a_j$ denote its unit purchasing
cost. We construct $a_j$ by dividing its historical
median transaction price by the training-set median
price and clipping the ratio to $[0.25,4]$.
The historical product price uses only transactions
preceding the forecast origin.
Overage and shortage costs are
$h_j=0.5a_j$ and $b_j=2a_j$, respectively.

Let $s_j$ denote the preceding week's sales and define
the order increment
\[
    d_j=\max\{1,\lceil s_j/8\rceil\}.
\]
The permissible order quantities are
$\mathcal Q_j=\{0,d_j,\ldots,16d_j\}$.
Let $q_j^{\mathrm{ref}}$ be the smallest permissible
quantity at least $\max\{1,s_j\}$.
The purchasing budget is
$B=0.8\sum_j a_jq_j^{\mathrm{ref}}$.

For observed sales $c_j$ and order quantity $w_j$,
the inventory cost is
\begin{equation}
    F(w,c)
    =
    \sum_{j=1}^{16}
    \left[
        h_j(w_j-c_j)_+
        +
        b_j(c_j-w_j)_+
    \right],
    \label{eq:app_inventory_cost}
\end{equation}
where $(u)_+=\max\{u,0\}$.
The feasible set is
\begin{equation}
    \mathcal W(x)
    =
    \left\{
        w:
        w_j\in\mathcal Q_j,\quad
        \sum_{j=1}^{16}a_jw_j\leq B
    \right\}.
    \label{eq:app_inventory_feasible}
\end{equation}
The feasible set and inventory cost coefficients may depend
on the observed context $x$. We suppress this dependence
in our notation; for each instance, they remain fixed
across predicted and observed parameter values.

\paragraph{Equivalent binary formulation.}
Index the permissible quantities as $q_{jk}=kd_j$,
$k=0,\ldots,16$, and let $z_{jk}\in\{0,1\}$ indicate
whether quantity $q_{jk}$ is selected for product $j$.
The feasible set becomes
\begin{equation*}
    \mathcal Z(x)
    :=
    \left\{
        z\in\{0,1\}^{16\times17}:
        \sum_{k=0}^{16}z_{jk}=1\ \forall j,\quad
        \sum_{j=1}^{16}\sum_{k=0}^{16}
        a_jq_{jk}z_{jk}\leq B
    \right\}.
    \label{eq:app_inventory_binary_feasible}
\end{equation*}
For sales vector $c$, define
\[
    v_{jk}(c)
    :=
    h_j(q_{jk}-c_j)_+
    +
    b_j(c_j-q_{jk})_+.
\]
Writing $v(c)$ and $z$ as vectors, the inventory problem
is equivalently
\begin{equation*}
    z^\star(v(c))
    \in
    \arg\min_{z\in\mathcal Z(x)} v(c)^\top z,
    \label{eq:app_inventory_binary}
\end{equation*}
with order quantities
$w_j=\sum_{k=0}^{16}q_{jk}z_{jk}$.
We use a fixed lexicographic order-quantity tie-breaking
rule independent of the observed sales.

\paragraph{SPO+ reward.}
The binary formulation has a linear objective and a
feasible set independent of the unknown sales.
We therefore apply SPO+ to the predicted and observed
cost vectors $\hat v=v(\hat c)$ and $v=v(c)$:
\begin{equation}
    \ell_{\mathrm{SPO+}}(\hat v,v)
    :=
    \max_{z\in\mathcal Z(x)}
    (v-2\hat v)^\top z
    +
    (2\hat v-v)^\top z^\star(v).
    \label{eq:app_inventory_spoplus}
\end{equation}
The surrogate-reward conditions use
$-\ell_{\mathrm{SPO+}}(v(\hat c),v(c))$ as reward.
Evaluating Equation~\eqref{eq:app_inventory_spoplus}
requires an optimization call with coefficients
$2v(\hat c)-v(c)$, which generally differ from
$v(2\hat c-c)$ because the mapping from sales to
menu costs is nonlinear.

\subsection{Training Settings}
\label[appendix]{app:inventory_training}

\paragraph{Conventional predictors.}
Linear regression and XGBoost predict product-level sales.
Numerical features include seven daily sales lags,
their sum, historical price, observed product age,
calendar variables, and inventory parameters.
Product name, category, and color are one-hot encoded.

Descriptions are encoded using unigram and bigram
TF--IDF features.
Linear regression uses these features directly; XGBoost uses a
64-dimensional truncated-SVD representation.
All encoders and transformations are fitted on training
data. 

For linear regression, we search the regularization
strength $\alpha\in\{0.01,0.1,1,10,100,1000\}$.
Validation selects 
$\alpha=1000$.
For XGBoost, we search learning rates
$\{0.03,0.05,0.1\}$, maximum depths $\{3,6\}$, and
tree counts $\{100,300,600,1000\}$.
Validation selects a learning rate of 0.03,
depth six, and 100 trees.

\paragraph{Supervised fine-tuning.}
We fine-tune Qwen3.5-4B using rank-16 LoRA adapters.
The initial SFT run uses all 4,013 training instances
for one epoch, a batch size of 16, and an initial
learning rate of $10^{-4}$ with linear decay.
Cross-entropy is averaged over answer tokens within
each instance and then over instances; prompt tokens
are masked.
Validation evaluates updates 0, 125, and 251 and
selects update 125.

\paragraph{Policy optimization.}
The four non-thinking RL conditions share the SFT
initialization and training-instance schedule.
Each uses 64 updates with 16 instances and eight
responses per instance, totaling 8,192 training
responses per condition.
Learning-rate selection uses eight development updates
per candidate, followed by evaluation on the full
validation set. The final training runs start from the
shared SFT checkpoint with the selected learning rate.

Training uses temperature 1, top-$p=1$, and a maximum
of 256 response tokens.
We use one policy update per sampled batch and no
KL penalty.
Adam uses $(\beta_1,\beta_2)=(0.9,0.95)$,
$\epsilon=10^{-8}$, zero weight decay, and a
gradient-norm limit of one.

Checkpoints are selected at updates 0, 32, and 64.
The negative-SPO+ RLOO condition selects update 32;
the other three conditions select update 64.

\paragraph{Reasoning-trace initialization.}
The reasoning-trace SFT dataset contains 256 training
instances selected across strata defined by calendar
month, historical sales volume, and the number of
zero-sales targets.
GPT-6-luna receives each context and its observed-sales
target and generates a target-conditioned explanation.
The student is trained on the explanation followed by
the unchanged target JSON, with context tokens masked.
Reasoning-trace SFT uses rank-16 adapters, a batch size
of 16, and one epoch of 16 updates, with an initial
learning rate of $2\times10^{-5}$ and linear decay.
Subsequent RLOO training enables thinking and assigns
negative decision regret to the generated response
based on its final parsed sales prediction.

\section{Prompts}
\label[prompt]{app:prompts}

\subsection{Inventory Prediction}
\label[appendix]{app:inventory_prompts}

The predictor receives the system prompt below and a user
message containing the instance inputs as JSON.
These inputs include product information, historical sales,
and the downstream optimization problem, but exclude
the observed next-week sales.

\begin{aipromptbox}[System Prompt for Inventory Prediction]
\label{prompt:inventory_prediction}

\role{System}
\begin{Verbatim}[fontsize=\promptsize,breaklines]
You are a retail sales forecasting model.

TASK
Predict the total number of units sold for each product during
the forecast period, using the information provided in the input.

INPUT INTERPRETATION
- "origin" is the first day of the forecast period.
- "forecast_horizon_days" specifies the length of that period.
- "items" contains the products to forecast.
- "past_daily_sales" lists observed sales from oldest to newest
  over the seven days immediately preceding the origin.
- "past_seven_day_sales" is the sum of those historical daily sales.
- "age_days" is the number of days since the product's first
  observed purchase.
- Product descriptions, when provided, attributes, historical
  prices, and calendar information provide additional
  forecasting context.

OPTIMIZATION CONTEXT
The input includes an inventory optimization model explaining
how your forecasts will be used. A downstream solver will
choose order quantities using your forecasts, the inventory
costs, the permitted order quantities, and the purchasing budget.

Predict sales, not order quantities. Sales predictions do not
need to belong to the order menus or satisfy the purchasing
budget. Do not cap a forecast at the maximum permitted
order quantity.

OUTPUT REQUIREMENTS
- Return exactly one JSON object with exactly one key: "sales".
- The value of "sales" must be an array of nonnegative integers.
- Include exactly one prediction for each product in "items".
- Preserve the input product order: the first prediction
  corresponds to the first item, the second prediction to
  the second item, and so on.
- Predict total sales over the entire forecast period,
  not daily sales.
- Zero is a valid prediction. There is no fixed upper limit.
- Return only the JSON object. Do not include explanations,
  reasoning, Markdown, product identifiers, or additional keys.

OUTPUT FORMAT EXAMPLE
For an input containing 16 products, a correctly formatted
response is:
{"sales": [12, 0, 35, 8, 21, 4, 67, 2,
           15, 9, 0, 43, 6, 18, 3, 27]}

These are invented values illustrating the format only.
Generate predictions from the supplied input; do not copy
the example values. Each position in the array corresponds
to the product at the same position in "items".
\end{Verbatim}

\role{User}
\begin{Verbatim}[fontsize=\promptsize,breaklines]
[INSTANCE INPUT JSON]
\end{Verbatim}
\end{aipromptbox}

\subsection{Reasoning-Trace Generation}
\label[appendix]{app:reasoning_teacher_prompt}

To construct the reasoning-trace SFT examples, we provide
GPT-6-luna with a training context and its observed-sales
target using the following user prompt.
The placeholders \texttt{\{input\}} and \texttt{\{output\}}
are replaced by the instance inputs and observed-sales
JSON, respectively.
The generated explanation is paired with the unchanged
target JSON as the student's supervised response.
The observed-sales target is not included in the student's
input.

\begin{aipromptbox}[Teacher Prompt for Generating Reasoning Traces]
\label{prompt:reasoning_teacher}

\role{User}
\begin{Verbatim}[fontsize=\promptsize,breaklines]
You are given an input and a target output. Write an
evidence-based prediction explanation for that output,
using only the input as evidence.

Treat `{output}` as the fixed prediction to explain.
Every predicted value, category, and item association in
your explanation must match `{output}` exactly.
Do not revise or substitute predictions.

Begin immediately with a concrete observation from the input
and the inference it supports. Do not introduce your method,
describe what you are about to do, or open with a general
statement about the data. Avoid sentences such as "I use a
baseline," "Most estimates are uncertain," or "The predictions
should be interpreted cautiously." When a baseline is relevant,
incorporate its actual value directly into the comparison
rather than announcing the method.

For each prediction, or group of predictions with genuinely
similar evidence, identify concrete facts or patterns from
the input, explain their predictive implications, and connect
those implications to the estimate. Use specific numerical
evidence where available. Establish a simple input-derived
baseline when appropriate, and distinguish the predicted
direction and magnitude of change from that baseline.
Check arithmetic and comparisons: do not describe a value
below its baseline as an increase.

Give more attention to substantial departures from a baseline,
conflicting signals, and uncertain estimates. Group predictions
only when the same explanation applies; identify meaningful
exceptions. A broad statement about a pattern is not enough
to justify several different numerical estimates.

Write in a prospective style, such as: "Since [specific
evidence], I expect [implication relative to the baseline].
Assuming [necessary assumption], I predict [quantity or
outcome] to be [estimate]." Adapt this structure naturally
instead of repeating it mechanically. Incorporate output
values into their explanations; do not reproduce the complete
output as a vector, list, or separate answer.

Make each paragraph develop a prediction: connect evidence,
its implication, any necessary assumption, and the resulting
estimate. Prefer active phrasing such as "Since ..., I expect ..."
and "Although ..., I predict ..." over detached commentary
such as "the estimate is ..." or "the forecast is consistent
with ...". Place any qualification next to the specific
inference it limits. Do not add a generic opening, closing
summary, or standalone disclaimer, and do not use causal
connectors as a substitute for a supported connection.

Use the output to identify the predictions to explain,
not as evidence for them. Do not refer to the output as
supplied, ground truth, observed, or already known.
Do not claim that you made the prediction without access
to the output. Do not invent facts, causal relationships,
formulas, or numerical adjustments to force a match.
State assumptions as assumptions. If the evidence supports
only a direction or approximate magnitude, say so briefly;
if a departure is not supported by the input, acknowledge
that limitation rather than manufacturing a justification.
Make uncertainty specific to the prediction instead of
adding a generic disclaimer.

Return only the explanation in approximately 250-400 words,
starting with evidence and ending with a specific prediction
or its relevant qualification. Prioritize substantive
evidence-to-prediction connections over restating the task
or listing values.

INPUT:
{input}

OUTPUT:
{output}
\end{Verbatim}
\end{aipromptbox}

\end{document}